\documentclass{article} 
\usepackage{iclr2027_conference,times}

\usepackage{hyperref}
\usepackage{url}
\usepackage{booktabs}
\usepackage{graphicx}
\usepackage{adjustbox}
\usepackage{threeparttable}
\usepackage{subcaption}
\usepackage{multirow}
\usepackage{tabularx}
\usepackage{array}
\usepackage{amsmath} 
\usepackage{amssymb}
\usepackage{listings}

\usepackage{amsthm}
\theoremstyle{plain}

\lstdefinestyle{metadatacode}{
    basicstyle=\ttfamily\footnotesize,
    columns=fullflexible,
    keepspaces=true,
    showstringspaces=false,
    breaklines=true,
    frame=single
}

\usepackage{xcolor}

\title{Which Metadata Matters for LLM Tool Selection in MCP? Evidence from Controlled Factor-Level Interventions}

\author{Xiaofan Li$^{1}$, Fengchun Qiao$^{2}$, Xing Gao$^{1}$ \\
$^{1}$University of Delaware, Newark, DE, USA \\
$^{2}$University of South Florida, Tampa, FL, USA
}

\iclrfinalcopy 
\begin{document}

\maketitle
\fancyhead{}                          
\renewcommand{\headrulewidth}{0pt}

\begin{abstract}
In Model Context Protocol (MCP) systems, LLMs select tools from metadata rather than from their implementations.
Names and descriptions summarize what a tool does, while the parameter schema specifies the fields a tool accepts and how they are organized.
Our analysis of metadata for 367,174 MCP tools reveals substantial schema variation among functionally similar tools.
Many of these tools share names and similar descriptions but expose different parameter schemas.
This raises a key question: can parameter-schema differences influence tool selection before any call is made?
We study this question through controlled metadata interventions across five LLMs and five tasks.
Each comparison changes exactly one metadata factor (server name, tool name, description, or parameter schema) while keeping the others fixed.
We balance tool positions and use conditional logistic regression to estimate position-adjusted effects.
We find that: (1) parameter-schema changes produce the largest and most consistent selection shifts among the tested interventions.
Models tend to favor separate top-level parameters over grouped or composite forms.
(2) Descriptions that explain parameters also tend to increase selection, while name effects are weaker and less consistent.
(3) Targeted comparisons show that schema effects depend on the specific parameter names and on how the request represents the same information.
When only one schema satisfies an explicit request requirement, all five models select the compatible tool despite position preferences.
These results show that parameter schemas influence not only how tools are called, but also whether they are selected.
For MCP developers and evaluators, schema design and tool ordering can therefore affect which implementation is invoked even when functionality is unchanged.
\end{abstract}

\section{Introduction}
\label{sec:introduction}

\begin{figure*}[t]
    \centering
    \includegraphics[width=\textwidth]{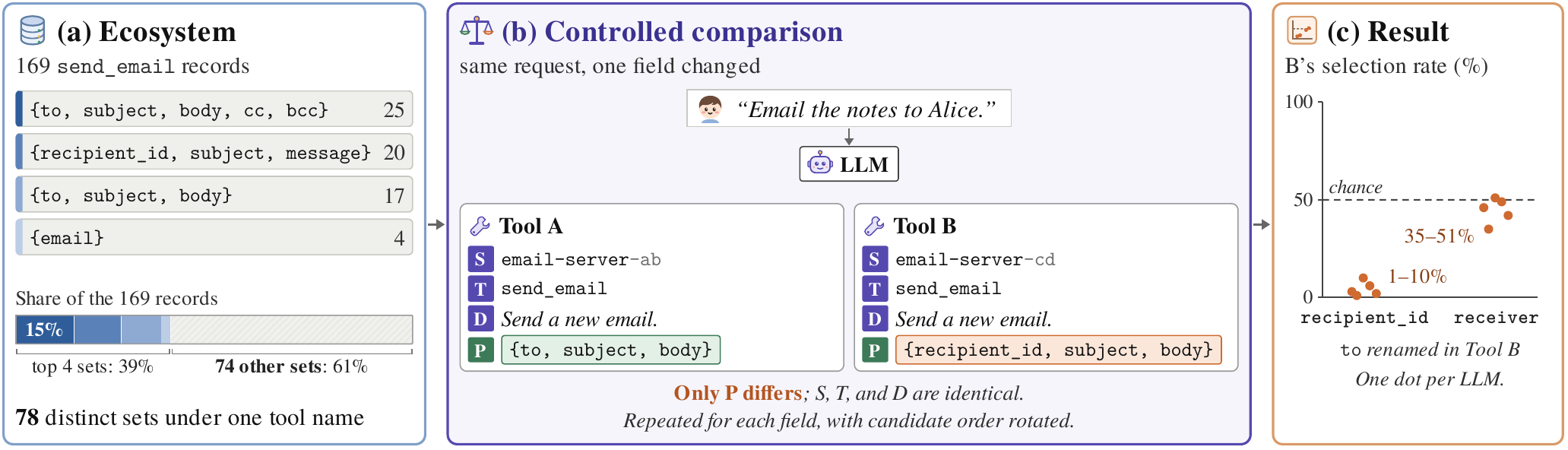}
   \caption{Overview of the study (full MCP workflow in Appendix Figure~\ref{fig:mcp_workflow}). 
(a) Tools that share a name often take different parameters: the 169 \texttt{send\_email} records in our dataset use 78 distinct parameter sets.
(b) In each comparison, the LLM receives the same request and candidate tools that differ in exactly one metadata field: server name (S), tool name (T), description (D), or parameters (P).
The host joins S and T into a single name field, and a two-letter server suffix only distinguishes the candidates.
Here, Tool B renames \texttt{to} to \texttt{recipient\_id}.
(c) Selection rate of Tool B in two-tool comparisons, with one dot per LLM.
Renaming \texttt{to} to \texttt{recipient\_id} lowers it to 1--10\%, whereas renaming it to \texttt{receiver} keeps it at 35--51\%, close to chance.}
    \label{fig:overview}
\end{figure*}

The Model Context Protocol (MCP) lets an application connect to tools from many independently developed MCP servers \citep{mcp_client,mcp_tools}.
As tool ecosystems grow, an LLM may need to choose among tools with overlapping functionality.
The model does not observe a tool's implementation directly; it sees an interface representation consisting of a server name, a tool name, a description, and a parameter schema (Figure~\ref{fig:overview}b).
Which tool the model selects determines which implementation runs, so the choice matters for the reliability of tool-using systems, for developers whose servers compete to be used, and for anyone who might edit tool metadata to steer the model.

These metadata factors are designed for different purposes.
Names and descriptions explicitly summarize what a tool does, whereas the parameter schema is primarily designed to specify how the tool should be invoked.
The MCP ecosystem shows why this distinction deserves attention.
In metadata for 367,174 MCP tools \citep{li2026first}, tools with the same function often share a name and a similar description but differ widely in their parameters: the 169 records named \texttt{send\_email} use 78 distinct parameter sets (Figure~\ref{fig:overview}a).
From their names and descriptions, these tools are easy to recognize as doing the same thing; they differ mainly in details intended for invocation, such as whether the recipient field is called \texttt{to} or \texttt{recipient\_id} (Appendix~\ref{app:send_email_cluster}).
This raises the possibility that such details already influence which tool is selected, before any call is made.
We ask whether they do, and how their influence compares with that of names and descriptions.

Prior work has shown that tool selection can be biased \citep{blankenstein2026biasbusters} and that edits to tool descriptions can shift which tool a model prefers \citep{faghih2025tool,wang2026mpma}; related work has also documented quality problems in MCP tool descriptions \citep{hasan2026model}.
The role of parameter schemas has received substantially less attention, particularly in controlled comparisons that isolate schema changes from changes to names and descriptions.

Ecosystem data alone cannot answer this question, because tools that differ in their schemas usually differ in other metadata as well.
We therefore intervene directly.
For five common tasks, we construct tool variants that differ in exactly one metadata field, using forms observed in the ecosystem, and present them together as candidates for the same user requests.
Because candidate order can itself affect selection, we balance tool positions and adjust for position with a conditional logit model \citep{train2003discrete}.

In this paper, we make three contributions.

\textbf{First}, we introduce a controlled framework for studying MCP tool selection.
It spans four metadata factors, five tasks, five LLMs, and candidate sets of two, three, and five tools.
Each comparison changes a single field, reuses the same requests, and balances candidate positions, isolating the effect of each tested edit.

\textbf{Second}, we find that schema edits produce the largest and most consistent shifts in selection among the tested edits.
We distinguish three kinds of schema edits: renaming parameters, grouping parameters into a single object, and splitting a composite input into separate factors.
Models tend to favor separate top-level fields, such as \texttt{owner} and \texttt{repo} rather than a single \texttt{owner/repo} string, and naming effects depend strongly on the particular names being compared.
Descriptions that explain parameters also tend to increase selection, whereas the effects of server and tool names are weaker and vary across models.
When candidates differ only in their names, models often pick the first-listed tool.

\textbf{Third}, we examine when and why schema effects arise.
Renaming \texttt{to} to \texttt{recipient\_id} reduces a tool's selection rate to 1--10\% across models, whereas renaming it to \texttt{receiver}, a word absent from the requests, leaves the tool competitive (35--51\%).
The effect therefore does not require word overlap with the request, and is consistent with \texttt{recipient\_id} signaling an identifier rather than an email address.
The preference for separate \texttt{owner} and \texttt{repo} fields weakens for every model, and reverses for one, when the same request gives the repository as a single \texttt{owner/repo} path.
Finally, when a request asks for a CC recipient that only one schema can accept, all five models select that tool, even when it is listed second.
Together, these results indicate that models use schema information to judge whether a tool fits a request during selection, rather than only to format arguments after selection.

These findings have practical consequences.
For server developers, schema design affects not only how a tool is called but whether it is called at all.
For host designers and evaluators, the strength of order effects means that tool-selection outcomes should be measured under varied candidate orders.

\section{Related Work}
\label{sec:background}

\paragraph{Tool selection.}
Most tool-use benchmarks evaluate whether a model selects and calls the correct tool~\citep{huang2024metatool,patil2025berkeley}, including in MCP settings~\citep{luo2025mcp,wang2026mcp} and after retrieval narrows the candidate set~\citep{gan2025rag}. Robustness studies perturb queries, toolkits, or tool documentation and measure the resulting loss in accuracy~\citep{ye2024rotbench,rabinovich2025robustness}. In these settings, one candidate is correct. We instead study selection among functionally similar tools, where no candidate is wrong, and the question is what drives the model's choice.

\paragraph{Metadata effects on selection.}
Selection among equivalent tools is biased~\citep{blankenstein2026biasbusters}, and edits to tool descriptions can increase a tool's usage by more than an order of magnitude~\citep{faghih2025tool}. Adversarial work exploits the same channel through optimized names and descriptions~\citep{sneh2025tooltweak,wang2026mpma}, injected tool documents~\citep{shi2025prompt}, and poisoned MCP metadata~\citep{wang2026mcptox,li2026first}. These studies focus on names and descriptions; parameter schemas are rarely varied in isolation. \citet{anand2026diagnosing} include a parameter trap that adds an unsatisfiable required argument, testing whether agents avoid unusable tools. In our comparisons, every candidate can serve the request; schemas differ only in naming, structuring, or decomposition, so a selection shift reflects preference rather than feasibility.

\paragraph{Tool documentation.}
Tool documentation alone can enable zero-shot tool use~\citep{hsieh2023tool}, and concise rewrites of tool instructions improve agent performance~\citep{yuan2025easytool}. For MCP, tool descriptions often contain quality defects, and augmenting them involves trade-offs between task performance and execution cost~\citep{hasan2026model}. These works mainly examine how documentation affects tool use once a tool is chosen; we study how tool metadata, from descriptions to parameter schemas, shapes which tool is chosen in the first place.

\paragraph{Order effects and choice modeling.}
LLMs are sensitive to option order in multiple-choice questions~\citep{zheng2024large,pezeshkpour2024large}, and AI shopping agents show strong position biases along with sensitivity to product descriptions~\citep{allouah2026your,kumar2024manipulating}. We therefore balance tool positions and include them as covariates in a conditional logit model~\citep{train2003discrete}, which estimates both metadata effects net of position and the position effects themselves.




\section{Modeling Tool Selection}
\label{sec:model}

We treat each trial as a choice among candidate tools and estimate how changing one metadata factor shifts that choice, separately from the effect of where each tool appears.

\subsection{Setup}

Given a task request $q$ and an ordered list of candidate tools $C=(c_1,\dots,c_m)$ with $m\in\{2,3,5\}$, we represent each tool by the four metadata factors the model observes:
\begin{equation}
c_i = (S_i, T_i, D_i, P_i),
\end{equation}
namely the server name, tool name, description, and parameter schema. Tool $c_i$ appears at position $i$, and $Y\in\{1,\dots,m\}$ denotes the position of the selected tool.

\paragraph{Metadata variants.}
For each factor $f\in\{S,T,D,P\}$, we define two variants, $v_{\text{high}}$ and $v_{\text{low}}$, and let $x_{i,f}\in\{0,1\}$ indicate whether tool $c_i$ uses $v_{\text{high}}$ ($x_{i,f}=1$) or $v_{\text{low}}$ ($x_{i,f}=0$). For server names, tool names, and parameter schemas, $v_{\text{high}}$ is the more common form within the task cluster and $v_{\text{low}}$ a less common one (Appendix~\ref{app:tasks}); parameter alternatives differ in naming, structuring, or decomposition (Table~\ref{tab:parameter_schema_variations}). For descriptions, $v_{\text{low}}$ briefly states the tool's function, and $v_{\text{high}}$ adds explanations of parameters or return values, following prior work on description quality~\citep{hasan2026model}. A positive effect of $v_{\text{high}}$ thus indicates a preference for the more common form for $S$, $T$, and $P$, and for the more complete description for $D$.

\paragraph{Controlled comparisons.}
Each comparison varies a single factor $f$ across the $m$ candidates and keeps the other three fixed, so candidates differ only in $x_{i,f}$ and position.

\subsection{Conditional Logit Model}

\begin{table}[t]
    \centering
    \scriptsize
    \begin{threeparttable}
        \caption{Parameter variations across the five tasks.}
        \label{tab:parameter_schema_variations}
        \renewcommand{\arraystretch}{1.15}
        \begin{tabularx}{\textwidth}{
            @{}>{\raggedright\arraybackslash}p{2.2cm}
            >{\raggedright\arraybackslash}p{2.8cm}
            >{\raggedright\arraybackslash}X
            >{\raggedright\arraybackslash}X@{}
        }
            \toprule
            \textbf{Dimension} & \textbf{Task} & \textbf{$v_{\text{high}}$} & \textbf{$v_{\text{low}}$} \\
            \midrule
            \multirow{2}{*}{Naming}
                & Send Email   & \texttt{to}       & \texttt{recipient\_id} \\
                & Get Weather  & \texttt{location} & \texttt{input} \\
            \midrule
            \multirow{2}{*}{Structuring}
                & Create Issue & \texttt{(owner, repo, ...)}  & \texttt{params\_dict} \\
                & Place Order  & \texttt{(customer\_id, ...)} & \texttt{create\_order\_requests} \\
            \midrule
            Decomposition
                & Create Pull Request & \texttt{(owner, repo)} & \texttt{repository} \\
            \bottomrule
        \end{tabularx}
        \begin{tablenotes}[flushleft]
            \scriptsize
            \item \textit{Note.} $v_{\text{high}}$ is the more common form in each task cluster. Naming changes parameter names; structuring groups separately declared parameters into a single container; decomposition splits a composite parameter into multiple parameters. Full parameter definitions are provided in Table~\ref{tab:metadata_variant_summary} in the appendix.
        \end{tablenotes}
    \end{threeparttable}
\end{table}

We model the selected position with a conditional logit model~\citep{train2003discrete}:
\begin{equation}
\Pr(Y=i \mid q, C) = \frac{\exp(\beta_f x_{i,f} + \gamma_i)}{\sum_{j=1}^{m}\exp(\beta_f x_{j,f} + \gamma_j)},
\label{eq:clogit}
\end{equation}
where $\beta_f$ measures the effect of using $v_{\text{high}}$ rather than $v_{\text{low}}$, and $\gamma_i$ measures the effect of appearing at position $i$ rather than last ($\gamma_m=0$). Because the other three factors are fixed across candidates, their contributions are constant within a trial and cancel in Eq.~\eqref{eq:clogit}. Estimated jointly, $\beta_f$ captures the metadata effect with position held fixed, and $\gamma_i$ captures the position effect with metadata held fixed; a positive $\beta_f$ thus reflects a preference for $v_{\text{high}}$ beyond any position preference.

\paragraph{Estimation.}
Each trial forms one choice group. For each factor, LLM, task, and candidate-set size, we estimate $(\beta_f,\gamma)$ by maximizing
\begin{equation}
\mathcal{L}(\beta_f,\gamma) = \sum_{r=1}^{R} \log \Pr(Y = Y_r \mid q_r, C_r),
\end{equation}
where $Y_r$, $q_r$, and $C_r$ are the selected position, request, and candidate list in trial $r$, and $R$ is the number of valid trials, i.e., trials in which the model calls a candidate tool (Section~\ref{sec:setup}).
We fit each model by maximum likelihood and report 95\% Wald confidence intervals, with standard errors from the inverse observed information. When a fit fails because choices are completely or quasi-completely separated, no finite estimate exists, and we mark the estimate as \emph{sep}.

\subsection{Selection Rates}

To describe observed choices without model assumptions, we also report selection rates. Let $H_r = x_{Y_r,f}$ indicate whether the tool selected in trial $r$ uses $v_{\text{high}}$. The $v_{\text{high}}$ selection rate and the position-$k$ selection rate are
\begin{equation}
\mathrm{SR}^{\text{high}}_f = \frac{1}{R}\sum_{r=1}^{R} H_r,
\qquad
\mathrm{SR}^{\text{pos}}_k = \frac{1}{R}\sum_{r=1}^{R} \mathbf{1}[Y_r = k],
\end{equation}
computed separately for each candidate composition (e.g., 1H+2L; Section~\ref{sec:setup}), with $R$ counting the trials in that composition. Under uniform choice, the baseline for $\mathrm{SR}^{\text{high}}_f$ is the fraction of candidates using $v_{\text{high}}$, and that for $\mathrm{SR}^{\text{pos}}_k$ is $1/m$. Selection rates summarize observed choices, whereas the conditional-logit coefficients separate metadata effects from position effects.

\section{Experimental Setup}
\label{sec:setup}

\paragraph{Tasks and queries.}
We study five tasks: Send Email, Get Weather, Create Issue, Place Order, and Create Pull Request, which cover the parameter naming, structuring, and decomposition differences in Table~\ref{tab:parameter_schema_variations}. The metadata variants are listed in Appendix~\ref{app:metadata_variants}. For each task, we use 50 natural-language queries generated with GPT-5.6 from the task's core inputs (e.g., recipient, subject, and body for Send Email) and manually checked. Queries vary in wording and inputs and never mention server or tool names. The same queries are reused across all factors and assignments. For Create Pull Request, the queries form 25 matched pairs that give the same request but specify the owner and repository either separately (\emph{split}) or as a combined \texttt{owner/repo} path (\emph{path}); we analyze the two formats separately.
 
\paragraph{Models.}
We evaluate GPT-5.4 mini~\citep{openai2026gpt54mini} (GPT), Claude Sonnet 4.6~\citep{anthropic2026sonnet46} (Claude), Gemini 3.5 Flash~\citep{google2026gemini35flash} (Gemini), DeepSeek V4 Pro~\citep{deepseek2026v4} (DeepSeek), and Qwen3.7-Plus~\citep{qwen2026qwen37plus} (Qwen). Temperature and other sampling settings are left at their API defaults. Thinking is disabled where supported; for Gemini, we set the thinking level to minimal. Tool choice is set to \emph{auto} with parallel tool calls disabled, and we record the called tool as the model's selection. All experiments were run in August and September 2026.
 
\paragraph{Candidate tools.}
Each comparison presents $m\in\{2,3,5\}$ candidate tools. The target factor takes $v_{\text{high}}$ or $v_{\text{low}}$ in each candidate, and the other three factors are fixed at their $v_{\text{low}}$ forms. Beyond the target factor, candidates differ only in a random two-letter server suffix (e.g., \texttt{-xl}). As in MCP hosts, the server and tool names are joined into a single name field (e.g., \texttt{message-server-xl\_\_send\_gmail}). Using H and L to denote $v_{\text{high}}$ and $v_{\text{low}}$, the candidate compositions are 1H+1L for two tools; 1H+2L and 2H+1L for three tools; and 2H+3L and 3H+2L for five tools.
 
\paragraph{Assignments and presentation order.}
An assignment specifies which candidates use $v_{\text{high}}$. We enumerate all assignments with one H among two candidates, one or two H among three, and two or three H among five, yielding 2, 6, and 20 assignments, so each candidate uses $v_{\text{high}}$ equally often. Presentation order is a cyclic rotation of the candidates whose offset depends on the query and assignment, so each candidate appears at each position equally often; the conditional logit model adjusts for any remaining position imbalance. Each query is run once per factor and assignment, giving 400, 1,200, and 4,000 trials per task--model pair for two, three, and five tools, with trial order shuffled using a fixed random seed. For Create Pull Request, trials are split equally between the two query formats. 
We exclude trials in which the request failed or no candidate tool was called; in every task--model configuration, these account for at most 1.3\% of trials.

\section{Results}

\subsection{Metadata Effects on Tool Selection}
\label{sec:main_effects}

Table~\ref{tab:metadata_effects_three_tools} reports conditional-logit coefficients for three-tool comparisons across five tasks and five models, and Appendix Figure~\ref{fig:create_issue_3tools_selection} shows the corresponding selection rates for Create Issue.

\begin{table}[t]
    \centering
    \begin{threeparttable}
        \caption{Conditional-logit estimates of metadata effects on tool selection with three candidate tools.}
        \label{tab:metadata_effects_three_tools}
         \scriptsize
        \renewcommand{\arraystretch}{1.08}
        \begin{tabular*}{\linewidth}{@{\extracolsep{\fill}}llrrrr@{}}
            \toprule
            \textbf{Task} & \textbf{Model} & \textbf{S} & \textbf{T} & \textbf{D} & \textbf{P} \\
            \midrule
            \multirow{5}{*}{Send Email}
            & GPT-5.4 mini      & 2.39 & 1.98 & 1.37 & \textbf{4.13} \\
            & Claude Sonnet 4.6 & 1.74$^{\dagger}$ & 1.24$^{\dagger}$ & 0.00$^{\dagger}$ & \textbf{4.56} \\
            & Gemini 3.5 Flash  & 1.54 & 1.62 & 0.55 & \textbf{7.88} \\
            & DeepSeek V4 Pro   & 0.01$^{\dagger}$ & $-$0.83$^{\dagger}$ & 2.49 & \textbf{3.03} \\
            & Qwen3.7-Plus      & 5.64$^{\dagger\ddagger}$ & $-$0.30$^{\dagger}$ & 5.64$^{\dagger\ddagger}$ & 11.77$^{\dagger}$ \\
            \midrule
            \multirow{5}{*}{Get Weather}
            & GPT-5.4 mini      & $-$0.89 & 0.38$^{\dagger}$ & 1.02 & \textbf{4.44} \\
            & Claude Sonnet 4.6 & 2.15 & 0.65$^{\dagger}$ & 1.73 & \textbf{7.87} \\
            & Gemini 3.5 Flash  & 2.26 & 0.14$^{\dagger}$ & 1.72 & \textbf{5.22} \\
            & DeepSeek V4 Pro   & 1.01 & $-$0.26$^{\dagger}$ & \textbf{2.49} & 1.78 \\
            & Qwen3.7-Plus      & 0.00$^{\dagger\ddagger}$ & $-$7.34$^{\dagger}$ & 1.80 & \textbf{2.16} \\
            \midrule
            \multirow{5}{*}{Create Issue}
            & GPT-5.4 mini      & \textbf{7.12} & 0.01$^{\dagger}$ & 2.35 & 16.23$^{\dagger}$ \\
            & Claude Sonnet 4.6 & 0.20$^{\dagger}$ & 0.16$^{\dagger}$ & 1.72 & \textbf{6.32} \\
            & Gemini 3.5 Flash  & $-$2.29 & 0.23$^{\dagger}$ & 2.36 & \textbf{sep}$^{\ast}$ \\
            & DeepSeek V4 Pro   & 2.81 & 0.12$^{\dagger}$ & 1.71 & \textbf{4.84} \\
            & Qwen3.7-Plus      & sep & 0.20$^{\dagger}$ & 4.74 & \textbf{sep}$^{\ast}$ \\
            \midrule
            \multirow{5}{*}{Place Order}
            & GPT-5.4 mini      & 0.43 & 0.54 & 1.35 & \textbf{4.30} \\
            & Claude Sonnet 4.6 & 7.29$^{\dagger}$ & 0.01$^{\dagger}$ & 10.65$^{\dagger}$ & 19.41$^{\dagger}$ \\
            & Gemini 3.5 Flash  & 0.03$^{\dagger}$ & $-$11.32$^{\dagger}$ & 1.98 & \textbf{2.51} \\
            & DeepSeek V4 Pro   & $-$0.51$^{\dagger}$ & $-$0.47$^{\dagger}$ & 0.01$^{\dagger}$ & $-$0.33$^{\dagger}$ \\
            & Qwen3.7-Plus      & 0.03$^{\dagger\ddagger}$ & 0.03$^{\dagger\ddagger}$ & 0.03$^{\dagger\ddagger}$ & 11.81$^{\dagger}$ \\
            \midrule
            \multirow{5}{*}{\shortstack[l]{Create Pull Request\\(\textit{split})}}
            & GPT-5.4 mini      & $-$0.23$^{\dagger}$ & sep & 0.15$^{\dagger}$ & \textbf{3.82} \\
            & Claude Sonnet 4.6 & 1.48 & 0.53$^{\dagger}$ & $-$0.47$^{\dagger}$ & \textbf{5.47} \\
            & Gemini 3.5 Flash  & 0.59 & 0.39$^{\dagger}$ & 1.43 & \textbf{3.07} \\
            & DeepSeek V4 Pro   & 1.14 & 0.78 & 0.40$^{\dagger}$ & \textbf{1.67} \\
            & Qwen3.7-Plus      & 1.42 & 0.42$^{\dagger}$ & 1.16 & \textbf{6.05} \\
            \midrule
            \multirow{5}{*}{\shortstack[l]{Create Pull Request\\(\textit{path})}}
            & GPT-5.4 mini      & 1.27 & $-$0.49$^{\dagger}$ & 0.50 & \textbf{1.95} \\
            & Claude Sonnet 4.6 & \textbf{3.08} & 0.89 & $-$0.07$^{\dagger}$ & 2.78 \\
            & Gemini 3.5 Flash  & 0.85 & 0.65 & 1.13 & \textbf{1.48} \\
            & DeepSeek V4 Pro   & \textbf{1.23} & 0.28$^{\dagger}$ & 0.72 & 0.51 \\
            & Qwen3.7-Plus      & 2.25 & 8.73$^{\dagger}$ & \textbf{2.71} & $-$1.01 \\
            \bottomrule
        \end{tabular*}
        \begin{tablenotes}[flushleft]
            \scriptsize
            \item \textit{Note.} Entries are conditional-logit coefficients ($\beta_f$) for $v_{\text{high}}$ vs.\ $v_{\text{low}}$, adjusted for position. S: server name; T: tool name; D: description; P: parameter schema. $^{\dagger}$95\% confidence interval includes zero; large estimates with $^{\dagger}$ (e.g., 16.23) reflect near-separation rather than weak effects. sep: separation, so no finite estimate exists; $^{\ast}$the model selected a $v_{\text{high}}$ tool in every trial. $^{\ddagger}$Qwen's choices in these comparisons are determined by position, so its metadata effects are not identified and these estimates should not be interpreted as effects. For Create Pull Request, \textit{split} and \textit{path} queries give the owner and repository separately or as a combined \texttt{owner/repo} path.
        \end{tablenotes}
    \end{threeparttable}
\end{table}


\paragraph{Parameter schemas have the largest and most consistent effects.}
Of the 28 numerical $P$ estimates, 23 have 95\% confidence intervals that exclude zero, and 22 of these are positive. The two remaining $P$ cells are separated because Gemini and Qwen select a $v_{\text{high}}$ tool in every Create Issue trial. Among coefficients whose intervals exclude zero, $P$ is the largest in 19 of the 30 task--model conditions, and its median magnitude (3.82) is more than twice that of $S$ (1.48). Several large estimates with wide intervals, such as GPT's 16.23 in Create Issue, reflect near-separation rather than weak effects. For the two naming tasks, Send Email and Get Weather, all five models have positive $P$ estimates, favoring the common names \texttt{to} and \texttt{location}; Section~\ref{sec:why} shows that such naming effects depend on the particular names being compared.

\paragraph{Models favor separate top-level fields.}
When parameters are grouped into a single object, models tend to prefer the variant with separate top-level parameters. In Create Issue, $v_{\text{high}}$ tools are selected in 93.3\%--100\% of trials across models and compositions, well above the corresponding baselines. In Place Order, $P$ estimates are positive for four models, while DeepSeek shows no clear preference ($\beta_P=-0.33$, interval including zero). For decomposition, all five models favor separate \texttt{owner} and \texttt{repo} parameters over a single \texttt{repository} field when the query gives the owner and repository separately, and four still do so when it gives a combined \texttt{owner/repo} path. Section~\ref{sec:why} examines how this preference changes with the request format. 

\paragraph{Descriptions that explain parameters increase selection.}
Of the 30 $D$ estimates, 21 have intervals that exclude zero, and all 21 are positive. This holds for every model in Get Weather and Create Issue, and for Gemini in every task and query format. In Create Issue, $v_{\text{high}}$ descriptions are also generally selected above their composition-specific baselines (Appendix Figure~\ref{fig:create_issue_3tools_selection}). Explaining parameters thus increases selection even when the brief description already states the tool's function.

\paragraph{Name effects are weaker, and position often decides.}
Tool names rarely matter: only 6 of 29 numerical $T$ estimates have intervals that exclude zero. Server-name effects are more frequent (19 of 29), but they are smaller than schema effects and reverse for two task--model pairs (GPT in Get Weather and Gemini in Create Issue). When candidates differ only in names, position often determines the choice. Under tool-name comparisons in Create Issue, $v_{\text{high}}$ selection rates stay near the uniform-choice baselines, whereas GPT, Claude, DeepSeek, and Qwen select the first-listed tool in 79\%--96\% of trials (Appendix Figure~\ref{fig:create_issue_3tools_selection}). Under server-name comparisons, Claude, DeepSeek, and Qwen show similar position preferences, whereas GPT favors $v_{\text{high}}$ and Gemini favors $v_{\text{low}}$.

\subsection{Candidate-Set Size and Position Effects}
\label{sec:set_size}

\begin{figure}[t]
    \centering
    \includegraphics[width=0.9\linewidth]
    {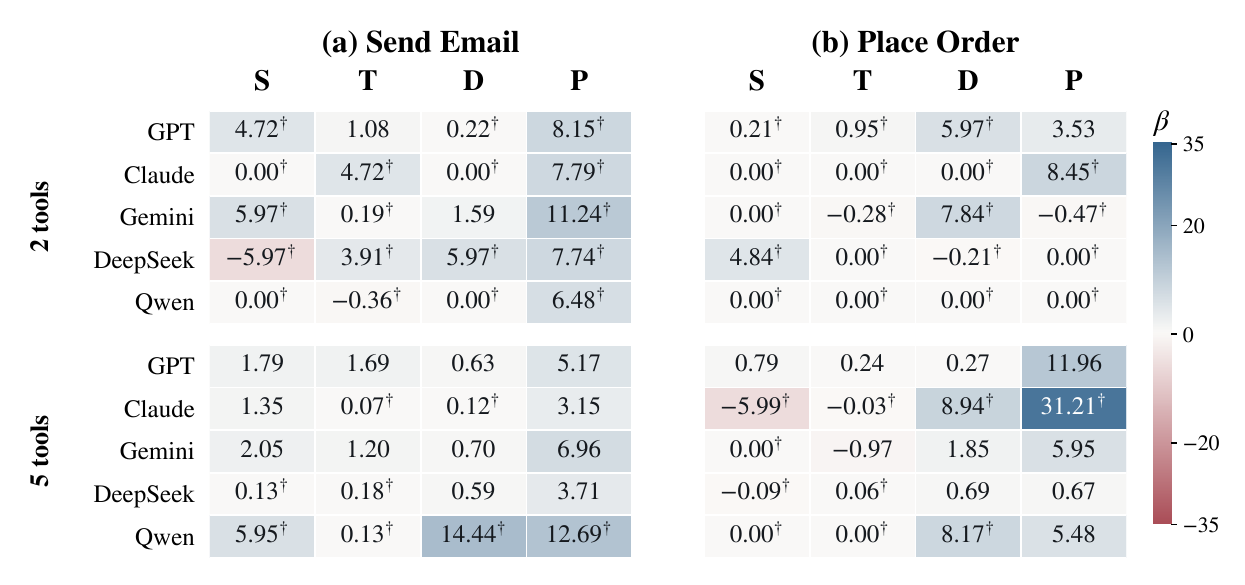}
    \caption{
        Metadata effects on tool selection with two and five candidate tools for (a) Send Email and (b) Place Order.
        Cells show conditional-logit coefficients ($\beta$) for $v_{\mathrm{high}}$ vs. $v_{\mathrm{low}}$, adjusted for presentation position.
        S: server name; T: tool name; D: description; P: parameter schema.
        Both panels use the same color scale. $\dagger$ indicates that the 95\% confidence interval
        includes zero.
    }
    \label{fig:candidate_set_size}
\end{figure}

\begin{figure}[t]
    \centering
    \includegraphics[width=0.9\linewidth]{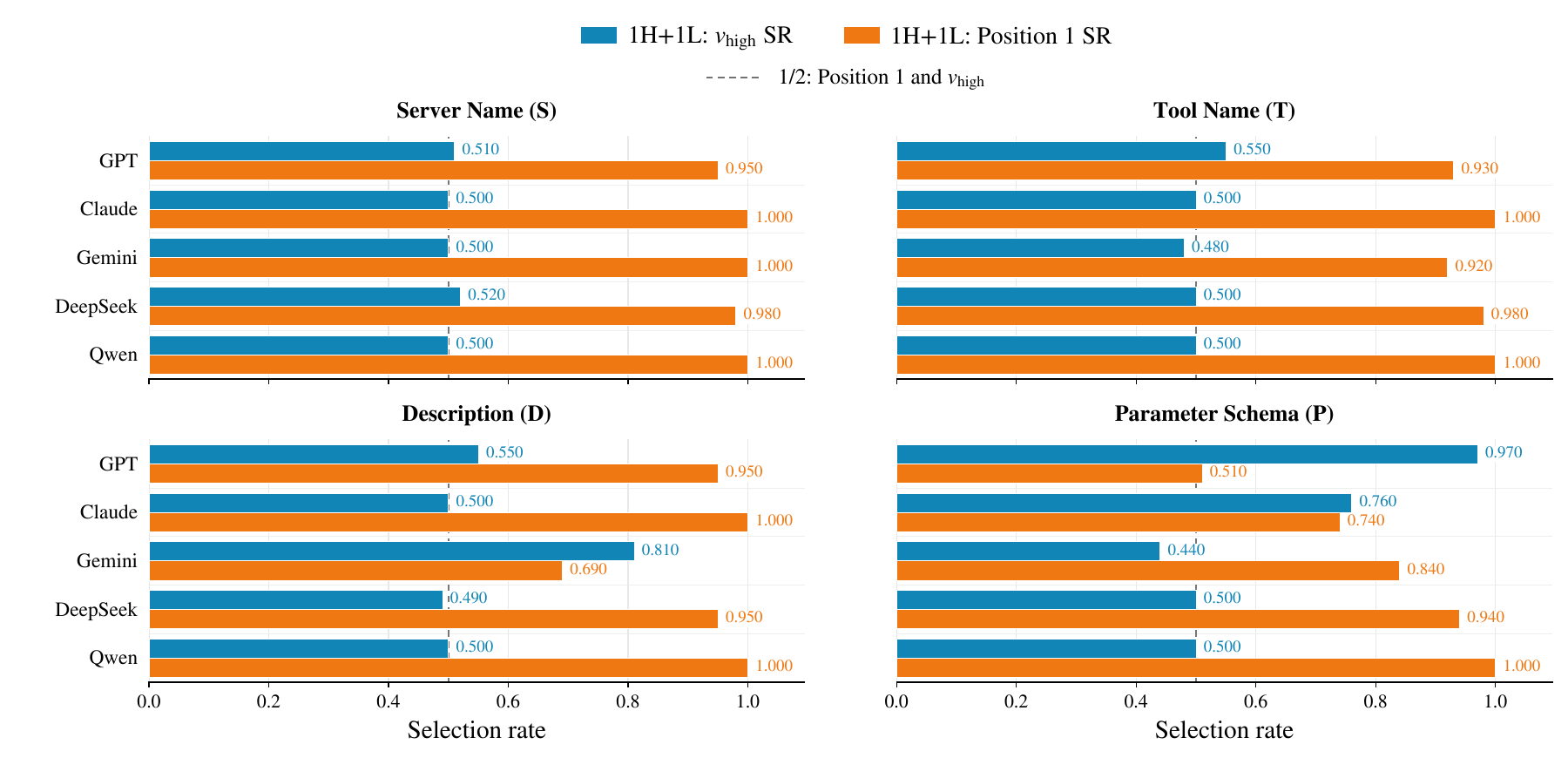}
    \caption{
        Two-tool selection rates for Place Order.
        Each comparison contains one $v_{\mathrm{high}}$ and one $v_{\mathrm{low}}$, with 100 valid trials per model and factor.
        Blue bars show the rate of selecting the $v_{\mathrm{high}}$ candidate; orange bars show the rate of selecting the first tool. The dashed line marks the uniform-choice baseline of 0.5 for both rates.
    }
    \label{fig:place_order_two_tool_sr}
\end{figure}

Figure~\ref{fig:candidate_set_size} compares coefficients with two and five candidate tools for Send Email and Place Order (see also Appendices~\ref{app:send_email_set_size}--\ref{app:send_email_composition}), and Figure~\ref{fig:place_order_two_tool_sr} shows the two-tool selection rates for Place Order.
 
 
\paragraph{Parameter effects persist with five tools.}
With five candidates, $P$ estimates are positive for all ten task--model pairs, and eight have intervals that exclude zero. The other two, Qwen in Send Email (12.69) and Claude in Place Order (31.21), are large estimates that reflect near-separation. In seven of the ten pairs, $P$ is the largest coefficient whose interval excludes zero. The schema preference found with three tools therefore does not depend on a particular candidate-set size.
 
\paragraph{With two tools, position dominates.}
Most two-tool estimates have intervals that include zero, and several are exactly zero. Figure~\ref{fig:place_order_two_tool_sr} explains why: under server- and tool-name comparisons in Place Order, all five models pick the first-listed tool in 92\%--100\% of trials, while $v_{\text{high}}$ rates stay near 50\%. Claude always picks the first tool under $S$, $T$, and $D$, and Qwen under all four factors, so Qwen's near-zero coefficients reflect separation by position rather than absent metadata effects.
 
\paragraph{Only strong metadata preferences overcome position.}
Parameter comparisons are the main exception: GPT selects the tool with separate top-level parameters in 97\% of trials but the first-listed tool in only 51\%, and Claude selects that tool in 76\%. Gemini instead follows descriptions (81\% $v_{\text{high}}$) rather than schemas (44\%). A single metadata difference thus outweighs position only when the preference is strong, so selection outcomes should be measured under varied candidate orders.

\subsection{Joint Effects of Metadata Factors}
\label{sec:factorial}

\begin{figure}[t]
    \centering
    \includegraphics[width=0.9\linewidth]{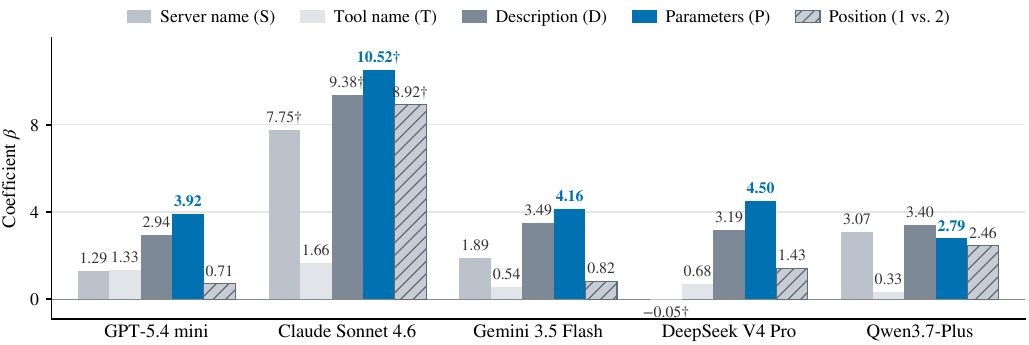}
    \caption{Joint metadata and position effects in the two-tool factorial experiment (Send Email). Bars show conditional-logit coefficients ($\beta$) for $v_{\text{high}}$ vs.\ $v_{\text{low}}$ in each field and for position 1 vs.\ 2, estimated jointly; parameter effects in blue. $\dagger$: 95\% confidence interval includes zero.}
    \label{fig:factorial-beta-grouped}
\end{figure}

The comparisons above vary one field at a time. To test whether schema effects persist when fields vary together, we run a two-tool factorial experiment on Send Email in which the two candidates differ in all four fields. For each field, one candidate uses $v_{\text{high}}$ and the other $v_{\text{low}}$; which candidate receives each variant is crossed across the four fields and presentation order, yielding $2^5=32$ conditions. Each of the 50 queries is run once per condition, giving 1,600 trials per model, and we estimate the effects of $S$, $T$, $D$, $P$, and position jointly (Figure~\ref{fig:factorial-beta-grouped}). The parameter contrast here concerns naming (Table~\ref{tab:parameter_schema_variations}).

\paragraph{Parameter effects remain the largest.}
$P$ has the largest coefficient for four of the five models. For GPT, Gemini, and DeepSeek, these coefficients (3.92, 4.16, and 4.50) have intervals that exclude zero; Claude's estimates are large for most factors but imprecise. Qwen is the exception, with its largest coefficient for descriptions ($\beta_D=3.40$), although its parameter effect remains positive ($\beta_P=2.79$). Descriptions rank second for GPT, Gemini, and DeepSeek, whereas tool-name effects stay small (at most 1.66), consistent with Section~\ref{sec:main_effects}.

\paragraph{Joint differences can outweigh position.}
For the four models with precise estimates, the position effect (0.71--2.46) is smaller than the parameter effect. Unlike the single-field comparisons in Section~\ref{sec:set_size}, where position often decides, choices among candidates that differ in several fields depend more on metadata than on position. Parameter-name effects thus persist when multiple metadata fields vary together.

\begin{figure}[t]
    \centering
    \includegraphics[width=\textwidth]{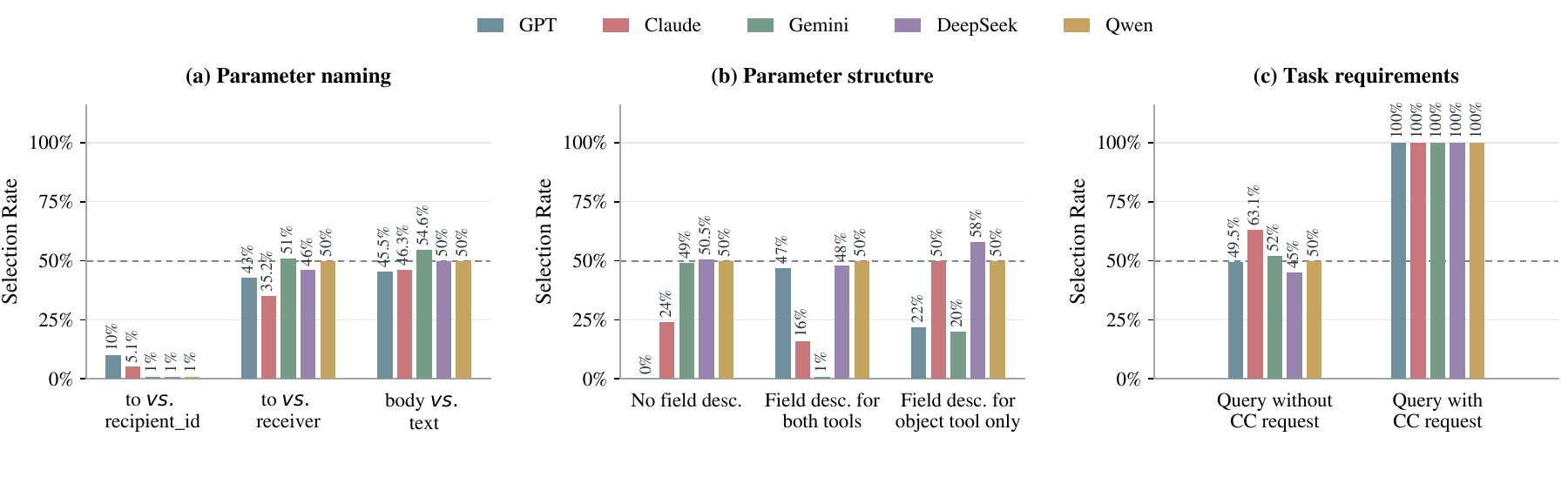}
    \caption{
        Two-tool selection rates under changes to parameter names, field descriptions, and task requirements.
        (a) Send Email: rate of selecting the tool with the renamed parameter; other parameters are unchanged.
        (b) Place Order: rate of selecting the tool with a single object parameter rather than separate named parameters.
        (c) Send Email: rate of selecting the tool with an additional optional \texttt{cc} parameter, with and without an explicit CC request.
        Dashed lines mark the 50\% baseline.
    }
    \label{fig:schema_properties}
\end{figure}

\subsection{When and Why Do Schema Effects Arise?}
\label{sec:why}


We now examine when and why schema effects arise, using two-tool comparisons that vary parameter names and task requirements (Figure~\ref{fig:schema_properties}; Appendices~\ref{app:naming_descriptions}--\ref{app:cc_naming}) and the matched query formats of Create Pull Request (Table~\ref{tab:metadata_effects_three_tools}).

\paragraph{Parameter names signal what input a tool expects.}
Replacing \texttt{to} with \texttt{recipient\_id} lowers the renamed tool's selection rate to 1.0\%--10.0\% across all five models, whereas replacing \texttt{to} with \texttt{receiver} keeps it at 35.2\%--51.0\%, and replacing \texttt{body} with \texttt{text} yields near-balanced rates of 45.5\%--54.6\% (Figure~\ref{fig:schema_properties}a). Because \texttt{receiver} never appears in the queries, a renamed tool can remain competitive without word overlap with the request. The drop for \texttt{recipient\_id} is instead consistent with the name signaling an identifier rather than an email address: \texttt{receiver} and \texttt{text} preserve the roles of recipient and message content, whereas \texttt{recipient\_id} suggests a different kind of input. Identical field descriptions remove this drop (Appendix~\ref{app:naming_descriptions}).

\paragraph{Schema preferences depend on how the request expresses inputs.}
The matched split and path queries for Create Pull Request differ only in whether the owner and repository are given separately or as a combined \texttt{owner/repo} path. Moving from split to path queries lowers the $P$ coefficient for every model: from 3.82 to 1.95 for GPT, 5.47 to 2.78 for Claude, 3.07 to 1.48 for Gemini, and 1.67 to 0.51 for DeepSeek. Qwen's coefficient reverses, from 6.05 to $-1.01$, so it favors the single \texttt{repository} field when the request supplies a single path. 
Field descriptions on structured parameters, by contrast, have no consistent effect: adding them to the object parameter in Place Order raises the object variant's selection rate for some models and lowers it for others (Figure~\ref{fig:schema_properties}b).


\paragraph{Task requirements can override position.}
In Send Email, one tool adds only an optional \texttt{cc} field to the shared \texttt{to}, \texttt{subject}, and \texttt{body}. Without a CC request, it is selected in 45.0\%--63.1\% of trials and the first-listed tool in 81.0\%--100\% (Figure~\ref{fig:schema_properties}c). With a CC request, all five models select it and satisfy the requirement in every valid trial, even when it is listed second. If it also renames \texttt{to} to \texttt{recipient\_id}, however, three models select it in only 53.0\%--67.4\% of these trials (Appendix~\ref{app:cc_naming}).

Together, these results indicate that models use schema information to judge whether a tool fits a request during selection, rather than only to format arguments after selection.


\section{Conclusion}
\label{sec:conclusion}

We studied what drives LLM tool selection among functionally similar MCP tools through controlled metadata interventions across five LLMs and five tasks.
After adjusting for position, parameter schemas had the largest and most consistent effects, favoring separate top-level fields across candidate-set sizes and under joint variation.
Descriptions that explain parameters also increased selection, whereas names mattered less and often yielded to position.
Schema effects depend on what parameter names signal and on how the request expresses its inputs, and explicit task requirements can override position.
Models thus use schemas to judge whether a tool fits a request during selection: schema design affects whether a tool is called at all, and tool-selection outcomes should be measured under varied candidate orders.

\paragraph{Limitations.}
Our findings are limited to the tested models, tasks, metadata variants, and candidate-set sizes, and to single-turn, LLM-generated English requests without tool execution.
We measure selection behavior rather than decision processes, so our interpretation of naming effects (e.g., \texttt{recipient\_id} signaling an identifier) remains behavioral.

\newpage

\section*{Ethics Statement}
This work studies how tool metadata affects LLM tool selection and involves no human subjects or personal data. The ecosystem analysis uses publicly available MCP tool metadata from \citet{li2026first}, and all queries were generated for this study with fictitious recipients and inputs (e.g., \texttt{example.com} addresses). Because our findings show that metadata changes can shift which tool is selected, they could inform attempts to steer model choices; we report them to help developers and host designers anticipate such effects and evaluate tool selection under varied metadata and candidate orders.

\section*{Reproducibility Statement}
Section~\ref{sec:model} specifies the conditional logit model and selection rates, and Section~\ref{sec:setup} describes the tasks, queries, models, API settings, candidate compositions, and trial schedules, including the dates of all API calls. Appendix~\ref{app:metadata_variants} lists the full metadata variants and parameter schemas. The supplementary material includes all queries and the request and analysis scripts, with the fixed random seeds used to generate trial schedules. Because the evaluated models are accessed through APIs that may change over time, exact results may differ in later runs.

\section*{AI Use Statement}
In this work, we used generative AI tools to identify functionally similar tools in the ecosystem data (Cursor with Grok 4.6; Appendix~\ref{app:clusters}), to generate the natural-language queries for each task (GPT-5.6; Section~\ref{sec:setup}), to help implement the experiment and analysis code, and to assist in summarizing results during writing. We did not use generative AI tools for any other purpose requiring disclosure, such as generating research ideas or formulating hypotheses. Additionally, we used generative AI tools to edit the paper for readability, identify relevant literature, and revise figures. We have reviewed all AI-assisted work: generated queries were checked for correctness and for the absence of server or tool names, AI-assisted code was tested by the authors, reported statistics were verified against the analysis outputs, and all cited works were checked against their original sources. The five LLMs evaluated in our experiments are the subjects of study rather than tools used to produce this work. We take responsibility for the final content of this work.



\newpage

\bibliography{ref}
\bibliographystyle{iclr2027_conference}

\clearpage
\appendix

\section{More Details on Tasks}
\label{app:tasks}

This appendix describes how we derive task clusters from the MCP ecosystem (\S\ref{app:clusters}), illustrates the variation within one cluster (\S\ref{app:send_email_cluster}), and details the queries and the targeted comparisons in Section~\ref{sec:why} (\S\ref{app:queries}--\S\ref{app:targeted}). The metadata variants and their coding are given in Appendix~\ref{app:metadata_variants}.

\subsection{Ecosystem Data and Task Clusters}
\label{app:clusters}

\paragraph{Metadata records.}
We use the metadata of 367,174 MCP tools collected by \citet{li2026first} from five MCP registries: mcp.so, MCP Market, MCP Store, Pulse MCP, and Smithery.
Each record contains the hosting platform, the server identifier, the tool name, the tool docstring, and the names of its parameters and return values. We map these fields to the four metadata factors: the server identifier (e.g., \texttt{cristip73/mcp-email-server}) to the server name $S$, the tool name (e.g., \texttt{send\_email}) to $T$, the docstring to the description $D$, and the parameter list (e.g., \texttt{to}, \texttt{subject}) to the parameter schema $P$.

\paragraph{Task clusters.}
As in \citet{blankenstein2026biasbusters}, we group functionally similar tools into task clusters. We use an LLM agent (Cursor with Grok 4.6 as the backbone model) to identify, based on tool descriptions, the tools that perform a given task, such as tools that can send an email. 
Because the data aggregate five registries, the same tool can appear more than once; we remove these duplicates, as well as tools whose parameter list or docstring is missing or empty. When counting name forms, we normalize tool names for case, underscores, hyphens, and spaces (e.g., \texttt{send\_email}, \texttt{sendEmail}, and \texttt{send-email} map to the same name).

\paragraph{From naming to structure and decomposition.}
Inspecting the resulting clusters, we found that tools sharing a name and a similar description often differ in their parameters, as in the Send Email cluster (Appendix~\ref{app:send_email_cluster}). The most direct difference is parameter naming, such as \texttt{to} versus \texttt{recipient\_id}. Starting from naming, we identified two further ways in which the same inputs can be specified: grouping separately declared parameters into a single container (structuring) and splitting a composite input into multiple parameters (decomposition). Table~\ref{tab:parameter_schema_variations} summarizes the three dimensions.

\paragraph{Task selection.}
For each dimension, we select representative task clusters that contain many tools and exhibit the corresponding variation: Send Email and Get Weather for naming, Create Issue and Place Order for structuring, and Create Pull Request for decomposition. After filtering, these clusters contain 288 (Send Email), 639 (Get Weather), 247 (Create Issue), 109 (Place Order), and 83 (Create Pull Request) records.
Within each cluster, we count the forms of server names, tool names, and parameter schemas to identify the more and less common variants (Appendix~\ref{app:metadata_variants}).

\subsection{Variation Within the Send Email Cluster}
\label{app:send_email_cluster}

\begin{figure}[ht]
    \centering
    \begin{minipage}[b]{0.3\linewidth}
        \centering
        \includegraphics[width=\linewidth]{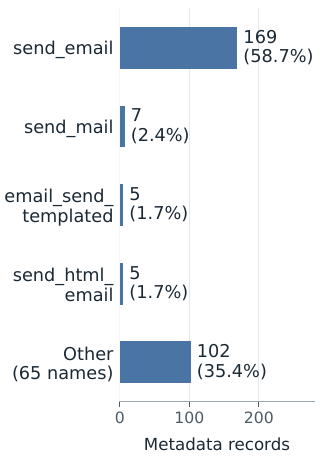}
        {\small (a) Tool name\par}
    \end{minipage}\hfill
    \begin{minipage}[b]{0.3\linewidth}
        \centering
        \includegraphics[width=\linewidth]{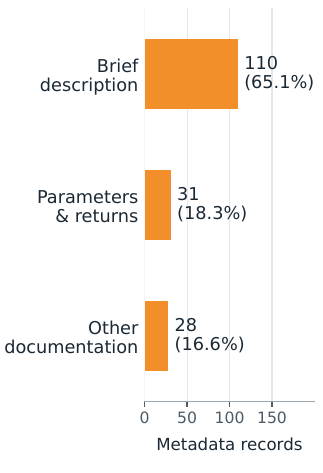}
        {\small (b) Description\par}
    \end{minipage}\hfill
    \begin{minipage}[b]{0.3\linewidth}
        \centering
        \includegraphics[width=\linewidth]{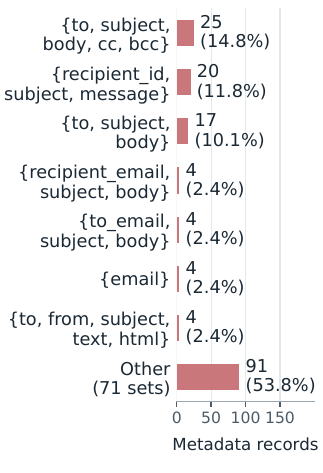}
        {\small (c) Parameters\par}
    \end{minipage}

    \caption{Tool names, description composition, and parameter sets
    in the Send Email cluster.}
    \label{fig:send_email_metadata}

    \begin{minipage}{\linewidth}
        \footnotesize\raggedright
        \textit{Notes.} Panel (a) uses all 288 deduplicated metadata records;
        tool names are normalized for case, underscores, hyphens, and spaces.
        Panels (b) and (c) use the 169 records mapped to \texttt{send\_email}.
        Parameter sets ignore order and preserve exact names.
        Low-frequency names and parameter sets remain counted in \emph{Other}.
    \end{minipage}
\end{figure}

Figure~\ref{fig:send_email_metadata} shows the distributions of tool names, description content, and parameter sets in the Send Email cluster. Among the 288 deduplicated records in the cluster, \texttt{send\_email} is the most common normalized name (169 records, 58.7\%), while 65 other names account for 35.4\%. Among the 169 \texttt{send\_email} records, most descriptions are brief (65.1\%), and 18.3\% explain parameters and return values. The same 169 records use 78 distinct parameter sets: the most common, \{\texttt{to}, \texttt{subject}, \texttt{body}, \texttt{cc}, \texttt{bcc}\}, covers only 14.8\% of records, and 71 low-frequency sets together account for 53.8\%. Tools that share a name and a similar description therefore differ widely in how their inputs are specified, which motivates isolating the parameter schema from the other fields.

\subsection{Queries}
\label{app:queries}

For each task, we prompted GPT-5.6 with the task's core inputs to generate 50 queries that vary in wording and values, and the authors checked each query for correctness and for the absence of server or tool names. 
All recipients and other personal details are fictitious. A representative Send Email query is:
\begin{quote}
\small
\textit{Email kevin@example.com with the subject 'Delivery Update' and tell him that the delivery is expected to arrive on Thursday.}
\end{quote}
For Create Pull Request, the queries are built from 25 base tasks, each specifying a repository, source and target branches, and a body, using five sentence templates. Each base task yields a matched pair that differs only in how the repository is referenced; branches, body, wording, tool assignments, presentation order, and shuffle seeds are otherwise identical. The repositories are illustrative and are never modified. For example:
\begin{quote}
\small
\textit{\textbf{Split:}} Create a new pull request in the GitHub repository `web-dashboard' under the organization `cedar-labs' to merge branch `fix/login-button' into branch `main'. Set its body to `Fix the unresponsive login button after valid credentials are entered.'. \\[2pt]
\textit{\textbf{Path:}} Create a new pull request in the GitHub repository `cedar-labs/web-dashboard' to merge branch `fix/login-button' into branch `main'. Set its body to `Fix the unresponsive login button after valid credentials are entered.'.
\end{quote}

\subsection{Targeted Comparisons}
\label{app:targeted}

The comparisons in Section~\ref{sec:why} use the two-tool design of Section~\ref{sec:setup}, with the two candidates differing only in the property under study.

\paragraph{Parameter names.} Starting from the Send Email tool with \{\texttt{to}, \texttt{subject}, \texttt{body}\}, we rename a single parameter in one candidate: \texttt{to} to \texttt{recipient\_id}, \texttt{to} to \texttt{receiver}, or \texttt{body} to \texttt{text}. All other parameters and fields are identical. We use the 50 Send Email queries with 200 trials per model per comparison. 

\paragraph{Field descriptions.} In Place Order, we compare the tool with separate top-level parameters (\texttt{customer\_id}, \texttt{item\_id}, \texttt{quantity}) against the tool that groups the same fields under a single \texttt{create\_order\_requests} object, under three conditions: no field descriptions, field descriptions on both tools, and field descriptions on the object tool only. Field descriptions are attached to the three leaf fields (``Numeric customer ID'', ``Numeric item ID'', and ``Quantity of items''); the wrapper object has no description, and both tools keep the tool-level description ``Place a new order''. We use the 50 Place Order queries, and each query is run once with each tool in the first position, giving 100 trials per condition and model.

\paragraph{Task requirements.} In Send Email, the base tool exposes \{\texttt{to}, \texttt{subject}, \texttt{body}\} and the extended tool adds only an optional \texttt{cc} field. Both tools are named \texttt{send\_email}, share the description ``Send a new email.'' without field descriptions, reject undeclared fields, and differ otherwise only in their two-letter server suffixes, which are counterbalanced across queries. We use 50 paired queries: in the \emph{unspecified} condition, the query asks to send an email without mentioning CC; in the \emph{required} condition, the same request, with the same recipient, additionally asks to CC a specified address. The paired queries are built on the Send Email queries of Appendix~\ref{app:queries} with the same generation and checking procedure.
Each query is run with the extended tool in each of the two positions, giving 100 trials per condition and 200 per model. Selection rates are computed over trials with exactly one call to a candidate tool. A trial satisfies the CC requirement if the model selects the extended tool with schema-valid arguments, the correct recipient, and the requested CC address.



\section{Metadata Variants and Coding}
\label{app:metadata_variants}

This appendix lists the two variants of each metadata factor for all five tasks. Table~\ref{tab:metadata_variant_summary} summarizes the names and parameter forms; Appendix~\ref{app:complete_metadata} gives the full descriptions and parameter schemas.

\subsection{Coding Criteria}
\label{app:coding_criteria}

\paragraph{Names and parameter forms.}
Within each task cluster (Appendix~\ref{app:clusters}), we code the more common form as $v_{\text{high}}$ ($x=1$) and a less common form as $v_{\text{low}}$ ($x=0$), counting both forms with the same rules. Server and tool names are counted after the normalization in Appendix~\ref{app:clusters}. For parameter schemas, the counting unit depends on the dimension: for naming, we count records that contain each core parameter name (e.g., \texttt{to} vs.\ \texttt{recipient\_id}); for structuring and decomposition, we count records with each structural form (separate top-level fields vs.\ a single container, or separate fields vs.\ a composite field). 
Counts of exact parameter sets are not used to order structural forms.

\paragraph{Descriptions.}
Descriptions are coded by completeness rather than frequency. The $v_{\text{low}}$ variant briefly states the tool's function; the $v_{\text{high}}$ variant adds explanations of parameters and, for Send Email and Get Weather, return values, following prior work on description quality~\citep{hasan2026model}.

\paragraph{Non-target factors.}
In all single-factor comparisons, non-target factors use their $v_{\text{low}}$ variants (Section~\ref{sec:setup}). When $P$ is varied, both candidates keep the brief description, so parameter names are not repeated in the description. When $D$ is varied, both candidates use the $v_{\text{low}}$ parameter schema, so the $v_{\text{high}}$ descriptions below refer to $v_{\text{low}}$ parameter names. In the factorial experiment (Section~\ref{sec:factorial}), each $v_{\text{high}}$ description instead lists the parameters of the schema assigned to that candidate.

\paragraph{Candidate identifiers.}
A stable two-letter suffix distinguishes each candidate's server name. In server-name comparisons, the suffix is held fixed while the base name changes. The exposed tool identifier joins the suffixed server name and the tool name with \texttt{\_\_} (e.g., \texttt{message-server-xl\_\_send\_gmail}). The suffix does not encode a metadata variant or a presentation position.

\subsection{Task-Specific Variants}
\label{app:task_specific_variants}

\begin{table}[!htbp]
\centering
\caption{Metadata variants for the five tasks. Server names are shown without the candidate-specific suffix. Description entries summarize the content; the full text appears in Appendix~\ref{app:complete_metadata}.}
\label{tab:metadata_variant_summary}
\small
\renewcommand{\arraystretch}{1.13}
\setlength{\tabcolsep}{4pt}
\begin{tabularx}{\linewidth}{@{}p{2.3cm}c>{\raggedright\arraybackslash}X>{\raggedright\arraybackslash}X@{}}
\toprule
\textbf{Task} & \textbf{Factor} & \textbf{$v_{\text{high}}$ (1)} & \textbf{$v_{\text{low}}$ (0)} \\
\midrule
Send Email & $S$ & \texttt{email-server} & \texttt{message-server} \\
 & $T$ & \texttt{send\_email} & \texttt{send\_gmail} \\
 & $D$ & Parameter and return explanations & Brief description \\
 & $P$ & \texttt{to, subject, body} & \texttt{recipient\_id, subject, message} \\
\midrule
Get Weather & $S$ & \texttt{weather-server} & \texttt{utility-server} \\
 & $T$ & \texttt{get\_weather} & \texttt{weather\_current} \\
 & $D$ & Parameter and return explanations & Brief description \\
 & $P$ & \texttt{location} & \texttt{input} \\
\midrule
Create Issue & $S$ & \texttt{github-server} & \texttt{pygithub-server} \\
 & $T$ & \texttt{create\_issue} & \texttt{create\_github\_issue} \\
 & $D$ & Parameter explanations & Brief description \\
 & $P$ & Separate parameters: \texttt{owner, repo, title, body} & One dictionary parameter: \texttt{params\_dict} \\
\midrule
Place Order & $S$ & \texttt{order-server} & \texttt{shop-server} \\
 & $T$ & \texttt{place\_order} & \texttt{create\_order} \\
 & $D$ & Parameter explanations & Brief description \\
 & $P$ & Separate parameters: \texttt{customer\_id, item\_id, quantity} & One object parameter: \texttt{create\_order\_requests} \\
\midrule
Create Pull Request & $S$ & \texttt{github-server} & \texttt{github-enterprise} \\
 & $T$ & \texttt{create\_pull\_request} & \texttt{create\_pr} \\
 & $D$ & Parameter explanations & Brief description \\
 & $P$ & \texttt{owner, repo, head, base, body} & \texttt{repository, body, base, head} \\
\bottomrule
\end{tabularx}
\end{table}


\clearpage
\subsection{Complete Descriptions and Parameter Schemas}
\label{app:complete_metadata}

The following listings reproduce the description wording, field types, required fields, and property order used in the experiments. As explained in Appendix~\ref{app:coding_criteria}, the $v_{\text{high}}$ descriptions refer to the $v_{\text{low}}$ parameter names.
\subsubsection{Send Email}
The parameter contrast changes \texttt{recipient\_id} to \texttt{to} and \texttt{message} to \texttt{body}. All three fields are required strings in both variants. The $v_{\text{high}}$ description also rewords the first line (``Send an email.'').

\noindent\textbf{Description, $v_{\text{low}}$.}
\begin{lstlisting}[style=metadatacode]
Send a new email.
\end{lstlisting}
\noindent\textbf{Description, $v_{\text{high}}$.}
\begin{lstlisting}[style=metadatacode]
Send an email.

Args:
    recipient_id: Recipient email address
    subject: Email subject
    message: Email body content

Returns:
    Confirmation of email sent
\end{lstlisting}
\noindent\begin{minipage}[t]{0.48\linewidth}
\vspace{0pt}
\textbf{Parameter schema, $v_{\text{low}}$.}
\begin{lstlisting}[style=metadatacode]
{
  "type": "object",
  "properties": {
    "recipient_id": {"type": "string"},
    "subject": {"type": "string"},
    "message": {"type": "string"}
  },
  "required": [
    "recipient_id",
    "subject",
    "message"
  ]
}
\end{lstlisting}
\end{minipage}
\hfill\begin{minipage}[t]{0.48\linewidth}
\vspace{0pt}
\textbf{Parameter schema, $v_{\text{high}}$.}
\begin{lstlisting}[style=metadatacode]
{
  "type": "object",
  "properties": {
    "to": {"type": "string"},
    "subject": {"type": "string"},
    "body": {"type": "string"}
  },
  "required": [
    "to",
    "subject",
    "body"
  ]
}
\end{lstlisting}
\end{minipage}

\clearpage
\subsubsection{Get Weather}
The parameter contrast changes the single required string field from \texttt{input} to \texttt{location}. 

\noindent\textbf{Description, $v_{\text{low}}$.}
\begin{lstlisting}[style=metadatacode]
Get current weather for a location.
\end{lstlisting}
\noindent\textbf{Description, $v_{\text{high}}$.}
\begin{lstlisting}[style=metadatacode]
Get current weather for a location.
Args:
    input (str): City name 
Returns:
    dict: WeatherAPI current weather JSON.
\end{lstlisting}
\noindent\begin{minipage}[t]{0.48\linewidth}
\vspace{0pt}
\textbf{Parameter schema, $v_{\text{low}}$.}
\begin{lstlisting}[style=metadatacode]
{
  "type": "object",
  "properties": {
    "input": {"type": "string"}
  },
  "required": [
    "input"
  ]
}
\end{lstlisting}
\end{minipage}
\hfill\begin{minipage}[t]{0.48\linewidth}
\vspace{0pt}
\textbf{Parameter schema, $v_{\text{high}}$.}
\begin{lstlisting}[style=metadatacode]
{
  "type": "object",
  "properties": {
    "location": {"type": "string"}
  },
  "required": [
    "location"
  ]
}
\end{lstlisting}
\end{minipage}

\clearpage
\subsubsection{Create Issue}
$v_{\text{low}}$ exposes an unrestricted dictionary parameter, \texttt{params\_dict}; its schema permits additional keys without enumerating the inner fields. $v_{\text{high}}$ explicitly declares four required string fields. The description of the dictionary explains the expected issue fields.

\noindent\textbf{Description, $v_{\text{low}}$.}
\begin{lstlisting}[style=metadatacode]
Create a new issue within a GitHub repository.
\end{lstlisting}
\noindent\textbf{Description, $v_{\text{high}}$.}
\begin{lstlisting}[style=metadatacode]
Create a new issue within a GitHub repository.

Args:
    params_dict (dict): Issue parameters containing these fields:
        owner (str): Repository owner's user or organization name
        repo (str): Repository name
        title (str): Issue title
        body (str): Issue body text
\end{lstlisting}
\noindent\begin{minipage}[t]{0.48\linewidth}
\vspace{0pt}
\textbf{Parameter schema, $v_{\text{low}}$.}
\begin{lstlisting}[style=metadatacode]
{
  "type": "object",
  "properties": {
    "params_dict": {
      "type": "object",
      "additionalProperties": true
    }
  },
  "required": [
    "params_dict"
  ]
}
\end{lstlisting}
\end{minipage}
\hfill\begin{minipage}[t]{0.48\linewidth}
\vspace{0pt}
\textbf{Parameter schema, $v_{\text{high}}$.}
\begin{lstlisting}[style=metadatacode]
{
  "type": "object",
  "properties": {
    "owner": {"type": "string"},
    "repo": {"type": "string"},
    "title": {"type": "string"},
    "body": {"type": "string"}
  },
  "required": [
    "owner",
    "repo",
    "title",
    "body"
  ]
}
\end{lstlisting}
\end{minipage}

\clearpage
\subsubsection{Place Order}
Both variants specify the same three required fields and types. $v_{\text{low}}$ groups them under \texttt{create\_order\_requests}; $v_{\text{high}}$ exposes them directly as top-level properties.

\noindent\textbf{Description, $v_{\text{low}}$.}
\begin{lstlisting}[style=metadatacode]
Place a new order
\end{lstlisting}
\noindent\textbf{Description, $v_{\text{high}}$.}
\begin{lstlisting}[style=metadatacode]
Place a new order

Args:
    create_order_requests (object): Order request containing these fields:
        customer_id (str): Numeric customer ID
        item_id (str): Numeric item ID
        quantity (int): Quantity of items
\end{lstlisting}
\noindent\begin{minipage}[t]{0.48\linewidth}
\vspace{0pt}
\textbf{Parameter schema, $v_{\text{low}}$.}
\begin{lstlisting}[style=metadatacode]
{
  "type": "object",
  "properties": {
    "create_order_requests": {
      "type": "object",
      "properties": {
        "customer_id": {"type": "string"},
        "item_id": {"type": "string"},
        "quantity": {"type": "integer"}
      },
      "required": [
        "customer_id",
        "item_id",
        "quantity"
      ]
    }
  },
  "required": [
    "create_order_requests"
  ]
}
\end{lstlisting}
\end{minipage}
\hfill\begin{minipage}[t]{0.48\linewidth}
\vspace{0pt}
\textbf{Parameter schema, $v_{\text{high}}$.}
\begin{lstlisting}[style=metadatacode]
{
  "type": "object",
  "properties": {
    "customer_id": {"type": "string"},
    "item_id": {"type": "string"},
    "quantity": {"type": "integer"}
  },
  "required": [
    "customer_id",
    "item_id",
    "quantity"
  ]
}
\end{lstlisting}
\end{minipage}

\clearpage
\subsubsection{Create Pull Request}
$v_{\text{low}}$ represents the repository as one \texttt{repository} string in \texttt{owner/repo} format. $v_{\text{high}}$ uses separate \texttt{owner} and \texttt{repo} strings. 

\noindent\textbf{Description, $v_{\text{low}}$.}
\begin{lstlisting}[style=metadatacode]
Create a new pull request
\end{lstlisting}
\noindent\textbf{Description, $v_{\text{high}}$.}
\begin{lstlisting}[style=metadatacode]
Create a new pull request

Args:
    repository (str): Repository identifier in owner/repo format
    body (str): Pull request body text
    base (str): Target branch to merge changes into
    head (str): Source branch containing the proposed changes
\end{lstlisting}
\noindent\begin{minipage}[t]{0.48\linewidth}
\vspace{0pt}
\textbf{Parameter schema, $v_{\text{low}}$.}
\begin{lstlisting}[style=metadatacode]
{
  "type": "object",
  "properties": {
    "repository": {"type": "string"},
    "body": {"type": "string"},
    "base": {"type": "string"},
    "head": {"type": "string"}
  },
  "required": [
    "repository",
    "body",
    "base",
    "head"
  ]
}
\end{lstlisting}
\end{minipage}
\hfill\begin{minipage}[t]{0.48\linewidth}
\vspace{0pt}
\textbf{Parameter schema, $v_{\text{high}}$.}
\begin{lstlisting}[style=metadatacode]
{
  "type": "object",
  "properties": {
    "owner": {"type": "string"},
    "repo": {"type": "string"},
    "head": {"type": "string"},
    "base": {"type": "string"},
    "body": {"type": "string"}
  },
  "required": [
    "owner",
    "repo",
    "head",
    "base",
    "body"
  ]
}
\end{lstlisting}
\end{minipage}

\section{Additional Results}
\label{app:additional_results}

\subsection{Three-Tool Selection Rates for Create Issue}
\label{app:create_issue_rates}

\begin{figure}[!htbp]
    \centering
    \includegraphics[width=\textwidth]{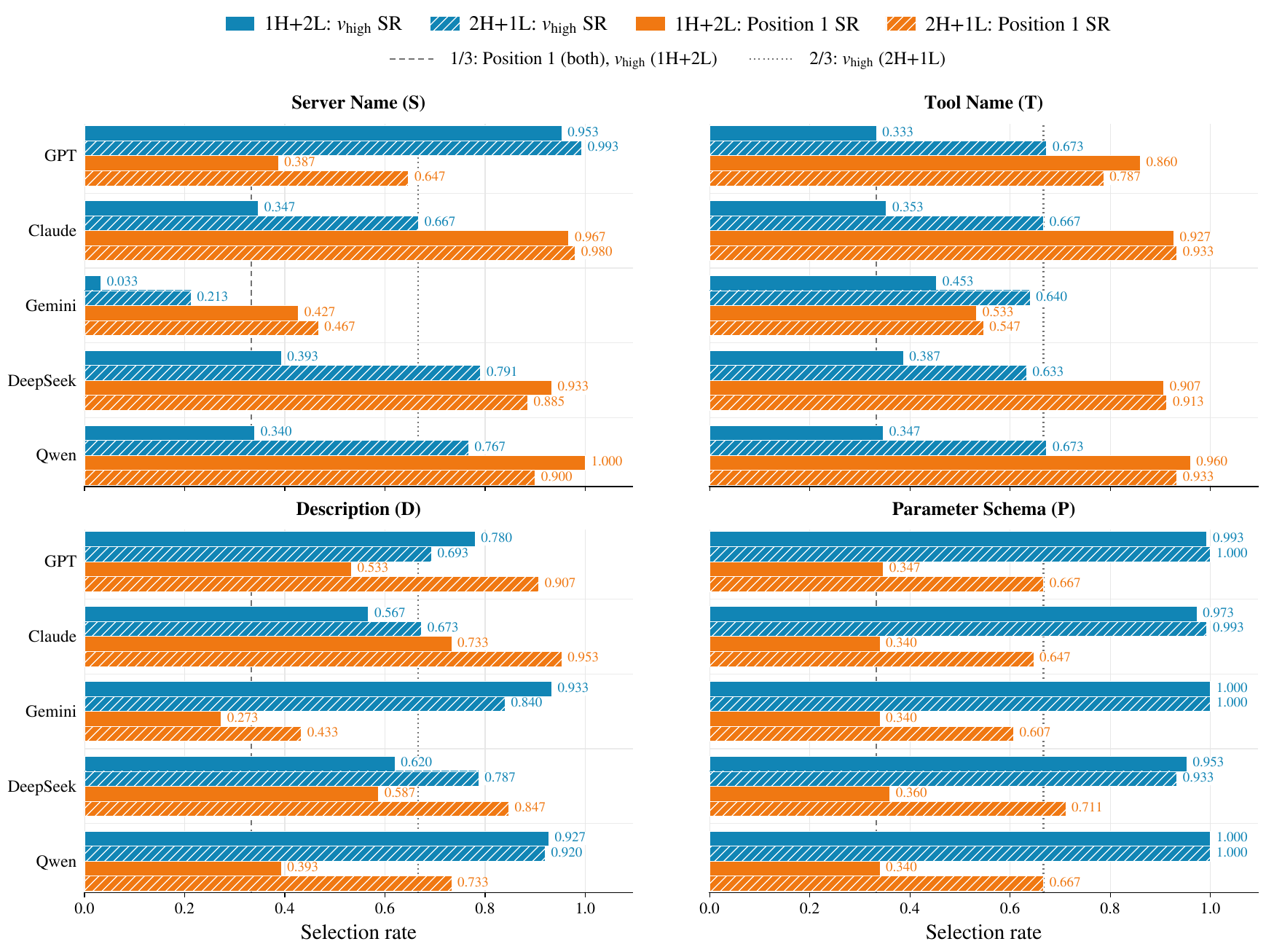}
    \caption{Observed three-tool selection rates for Create Issue across models and metadata factors. Blue bars show the $v_{\text{high}}$ selection rate and orange bars the position-1 selection rate; solid and hatched bars denote the 1H+2L and 2H+1L compositions, respectively. Each bar summarizes up to 150 trials. Under uniform choice, the $v_{\text{high}}$ baseline is $1/3$ for 1H+2L and $2/3$ for 2H+1L, and the position-1 baseline is $1/3$ for both.}
    \label{fig:create_issue_3tools_selection}
\end{figure}

Figure~\ref{fig:create_issue_3tools_selection} reports the observed selection rates behind the Create Issue estimates in Table~\ref{tab:metadata_effects_three_tools}, using the definitions in Section~\ref{sec:model}. Each factor shows a distinct pattern.

\paragraph{Parameter schemas.} All five models select a $v_{\text{high}}$ tool, which exposes separate top-level parameters, in 93.3\%--100\% of trials in both compositions. Gemini and Qwen do so in every trial, which is why their $P$ estimates are separated.

\paragraph{Descriptions.} $v_{\text{high}}$ selection rates are generally above the composition-specific baselines, most clearly for Gemini (93.3\% and 84.0\%) and Qwen (92.7\% and 92.0\%).

\paragraph{Tool names.} $v_{\text{high}}$ selection rates stay near the baselines (33.3\%--45.3\% for 1H+2L and 63.3\%--67.3\% for 2H+1L), whereas GPT, Claude, DeepSeek, and Qwen select the first-listed tool in 78.7\%--96.0\% of trials.

\paragraph{Server names.} Claude, DeepSeek, and Qwen again select the first-listed tool in most trials (88.5\%--100\%). GPT instead favors the $v_{\text{high}}$ server name (95.3\% and 99.3\%), and Gemini the $v_{\text{low}}$ one (3.3\% and 21.3\% $v_{\text{high}}$).

\subsection{Send Email Coefficients Across Candidate-Set Sizes}
\label{app:send_email_set_size}
 
Table~\ref{tab:metadata-beta} reports the Send Email coefficients with two, three, and five candidate tools. The two- and five-tool rows correspond to Figure~\ref{fig:candidate_set_size}a, and the three-tool rows repeat Table~\ref{tab:metadata_effects_three_tools}.
 
\paragraph{Parameter schemas.}
$P$ is positive in all 15 model--size combinations and is the largest estimate in every row except Qwen with five tools. With three and five tools, $P$ excludes zero for GPT, Claude, Gemini, and DeepSeek and is the largest estimate in each of these eight rows. The other $P$ estimates, all two-tool estimates (6.48--11.24) and Qwen's with three and five tools (11.77 and 12.69), are large but have intervals that include zero, reflecting near-separation rather than weak effects. The effect does not change uniformly with the number of candidates: from three to five tools, $P$ increases for GPT and DeepSeek but decreases for Claude and Gemini.
 
\paragraph{Descriptions and names.}
$D$ excludes zero for GPT, Gemini, and DeepSeek with three and five tools (0.55--2.49) and for Gemini with two tools (1.59), and it is always smaller than $P$ in these rows. Claude's $D$ never excludes zero. $T$ excludes zero only for GPT at all three set sizes (1.08--1.98) and for Gemini with three and five tools (1.62 and 1.20). $S$ excludes zero only for GPT and Gemini with three and five tools and for Claude with five tools (1.35--2.39). None of DeepSeek's or Qwen's name estimates excludes zero.
 
\paragraph{Two candidate tools.}
Only 2 of the 20 two-tool estimates exclude zero: GPT's $T$ (1.08) and Gemini's $D$ (1.59). Claude's and Qwen's $S$ and $D$ estimates are exactly zero, which occurs when choices are fully determined by position, and several others, such as DeepSeek's $S$ ($-$5.97) and $D$ (5.97), are large with intervals that include zero. With two tools, position therefore masks most metadata effects in Send Email, as in Place Order (Section~\ref{sec:set_size}).
 
\begin{table}[!htbp]
    \centering
    \begin{threeparttable}
        \caption{Conditional-logit estimates of metadata effects for Send Email with two, three, and five candidate tools.}
        \label{tab:metadata-beta}
        \footnotesize
        \renewcommand{\arraystretch}{1.05}
        \begin{tabular*}{\linewidth}{@{\extracolsep{\fill}}lcrrrr@{}}
            \toprule
            \textbf{Model} & \textbf{Tools} & \textbf{S} & \textbf{T} & \textbf{D} & \textbf{P} \\
            \midrule
            \multirow{3}{*}{GPT-5.4 mini}
            & 2 & 4.72$^{\dagger}$ & \textbf{1.08} & 0.22$^{\dagger}$ & 8.15$^{\dagger}$ \\
            & 3 & 2.39 & 1.98 & 1.37 & \textbf{4.13} \\
            & 5 & 1.79 & 1.69 & 0.63 & \textbf{5.17} \\
            \midrule
            \multirow{3}{*}{Claude Sonnet 4.6}
            & 2 & 0.00$^{\dagger}$ & 4.72$^{\dagger}$ & 0.00$^{\dagger}$ & 7.79$^{\dagger}$ \\
            & 3 & 1.74$^{\dagger}$ & 1.24$^{\dagger}$ & 0.00$^{\dagger}$ & \textbf{4.56} \\
            & 5 & 1.35 & 0.07$^{\dagger}$ & 0.12$^{\dagger}$ & \textbf{3.15} \\
            \midrule
            \multirow{3}{*}{Gemini 3.5 Flash}
            & 2 & 5.97$^{\dagger}$ & 0.19$^{\dagger}$ & \textbf{1.59} & 11.24$^{\dagger}$ \\
            & 3 & 1.54 & 1.62 & 0.55 & \textbf{7.88} \\
            & 5 & 2.05 & 1.20 & 0.70 & \textbf{6.96} \\
            \midrule
            \multirow{3}{*}{DeepSeek V4 Pro}
            & 2 & $-$5.97$^{\dagger}$ & 3.91$^{\dagger}$ & 5.97$^{\dagger}$ & 7.74$^{\dagger}$ \\
            & 3 & 0.01$^{\dagger}$ & $-$0.83$^{\dagger}$ & 2.49 & \textbf{3.03} \\
            & 5 & 0.13$^{\dagger}$ & 0.18$^{\dagger}$ & 0.59 & \textbf{3.71} \\
            \midrule
            \multirow{3}{*}{Qwen3.7-Plus}
            & 2 & 0.00$^{\dagger}$ & $-$0.36$^{\dagger}$ & 0.00$^{\dagger}$ & 6.48$^{\dagger}$ \\
            & 3 & 5.64$^{\dagger\ddagger}$ & $-$0.30$^{\dagger}$ & 5.64$^{\dagger\ddagger}$ & 11.77$^{\dagger}$ \\
            & 5 & 5.95$^{\dagger}$ & 0.13$^{\dagger}$ & 14.44$^{\dagger}$ & 12.69$^{\dagger}$ \\
            \bottomrule
        \end{tabular*}
        \begin{tablenotes}[flushleft]
            \footnotesize
            \item \textit{Note.} Entries are conditional-logit coefficients ($\beta_f$) for $v_{\text{high}}$ vs.\ $v_{\text{low}}$, adjusted for position. S: server name; T: tool name; D: description; P: parameter schema. $^{\dagger}$95\% confidence interval includes zero; large estimates with $^{\dagger}$ reflect near-separation rather than weak effects. $^{\ddagger}$As in Table~\ref{tab:metadata_effects_three_tools}, Qwen's choices are determined by position, so these estimates should not be interpreted as effects. Bold marks the largest estimate in each row among those whose interval excludes zero.
        \end{tablenotes}
    \end{threeparttable}
\end{table}

\subsection{Send Email Coefficients by Candidate Composition}
\label{app:send_email_composition}
 
Table~\ref{tab:composition-beta} splits the three- and five-tool Send Email estimates in Table~\ref{tab:metadata-beta} by candidate composition, that is, by how many candidates use $v_{\text{high}}$ (Section~\ref{sec:setup}). Because $\beta_f$ measures the preference for a $v_{\text{high}}$ candidate over a $v_{\text{low}}$ one, it can be compared across compositions with different uniform-choice baselines. We exclude Qwen from the comparisons below: none of its estimates excludes zero, and its three-tool $S$ and $D$ choices are determined by position.
 
\paragraph{Parameter schemas.}
Every numerical $P$ estimate is positive. For GPT, Claude, Gemini, and DeepSeek, $P$ excludes zero in 15 of the 16 multi-tool compositions; the exception, Gemini under 1H+2L (14.03), reflects near-separation. $P$ is also the largest of the four estimates in each of these 16 model--composition pairs except DeepSeek under 1H+2L, where $D$ (3.86) and $P$ (3.85) are nearly equal. The schema preference therefore holds whether $v_{\text{high}}$ schemas are in the minority or the majority.
 
\paragraph{Descriptions.}
Description effects depend on composition. For GPT, Gemini, and DeepSeek, $D$ is positive and excludes zero whenever $v_{\text{high}}$ descriptions are in the minority (1H+2L: 2.17--3.86; 2H+3L: 1.15--2.12). When they are in the majority (2H+1L and 3H+2L), these estimates are near zero or negative, and Gemini favors the $v_{\text{low}}$ description ($-$1.83 and $-$0.55, both excluding zero). A more complete description thus helps a tool when few candidates have one, but not when most do. Pooling over compositions masks this pattern: Gemini's three-tool estimate of 0.55 combines 3.50 and $-$1.83. Claude's $D$ never excludes zero.
 
\paragraph{Names.}
Server-name estimates are larger when most candidates use the $v_{\text{high}}$ server name, for GPT and Claude at both set sizes and for Gemini and DeepSeek with five tools (e.g., Claude: 0.00 vs.\ 2.88 with three tools and 0.45 vs.\ 2.08 with five). DeepSeek's five-tool $S$ reverses sign, from $-$0.78 under 2H+3L to 1.01 under 3H+2L, both excluding zero, so its pooled estimate (0.13) does not. Tool-name effects are concentrated in GPT and Gemini: $T$ excludes zero for GPT in every composition (1.08--2.51) and for Gemini in three of the four multi-tool compositions (0.99--3.34), but for DeepSeek only under 3H+2L (0.59) and never for Claude.
 
Overall, schema effects are the largest and most stable across compositions, whereas description and server-name effects depend on how the variants are distributed among the candidates.
 
\begin{table}[!htbp]
\centering
\scriptsize
\setlength{\tabcolsep}{3pt}
\renewcommand{\arraystretch}{1.12}
\captionsetup{skip=6pt}
\captionsetup[subtable]{font=footnotesize,skip=3pt}
\caption{Conditional-logit estimates of metadata effects for Send Email by candidate composition.}
\label{tab:composition-beta}
\begin{subtable}[t]{0.485\textwidth}
\centering
\caption{Server name (S)}
\begin{tabular*}{\linewidth}{@{\extracolsep{\fill}}lccccc@{}}
\toprule
\multirow{2}{*}{\textbf{Model}} & \textbf{2 Tools} & \multicolumn{2}{c}{\textbf{3 Tools}} & \multicolumn{2}{c}{\textbf{5 Tools}} \\
\cmidrule(lr){2-2}\cmidrule(lr){3-4}\cmidrule(lr){5-6}
 & \textbf{1H+1L} & \textbf{1H+2L} & \textbf{2H+1L} & \textbf{2H+3L} & \textbf{3H+2L} \\
\midrule
GPT-5.4 mini      & 4.72$^{\dagger}$ & 1.99 & 3.02 & 1.41 & 2.45 \\
Claude Sonnet 4.6 & 0.00$^{\dagger}$ & 0.00$^{\dagger}$ & 2.88 & 0.45$^{\dagger}$ & 2.08 \\
Gemini 3.5 Flash  & 5.97$^{\dagger}$ & 1.53$^{\dagger}$ & 1.55$^{\dagger}$ & 1.74 & 2.45 \\
DeepSeek V4 Pro   & $-$5.97$^{\dagger}$ & 0.01$^{\dagger}$ & 0.00$^{\dagger}$ & $-$0.78 & 1.01 \\
Qwen3.7-Plus      & 0.00$^{\dagger}$ & 7.10$^{\dagger\ddagger}$ & 14.34$^{\dagger\ddagger}$ & 6.26$^{\dagger}$ & 12.55$^{\dagger}$ \\
\bottomrule
\end{tabular*}
\end{subtable}
\hfill
\begin{subtable}[t]{0.485\textwidth}
\centering
\caption{Tool name (T)}
\begin{tabular*}{\linewidth}{@{\extracolsep{\fill}}lccccc@{}}
\toprule
\multirow{2}{*}{\textbf{Model}} & \textbf{2 Tools} & \multicolumn{2}{c}{\textbf{3 Tools}} & \multicolumn{2}{c}{\textbf{5 Tools}} \\
\cmidrule(lr){2-2}\cmidrule(lr){3-4}\cmidrule(lr){5-6}
 & \textbf{1H+1L} & \textbf{1H+2L} & \textbf{2H+1L} & \textbf{2H+3L} & \textbf{3H+2L} \\
\midrule
GPT-5.4 mini      & 1.08 & 2.51 & 1.21 & 1.78 & 1.55 \\
Claude Sonnet 4.6 & 4.72$^{\dagger}$ & 1.21$^{\dagger}$ & 1.27$^{\dagger}$ & $-$0.21$^{\dagger}$ & 0.35$^{\dagger}$ \\
Gemini 3.5 Flash  & 0.19$^{\dagger}$ & 3.34 & 0.16$^{\dagger}$ & 1.36 & 0.99 \\
DeepSeek V4 Pro   & 3.91$^{\dagger}$ & 0.00$^{\dagger}$ & $-$1.55$^{\dagger}$ & $-$0.24$^{\dagger}$ & 0.59 \\
Qwen3.7-Plus      & $-$0.36$^{\dagger}$ & 0.78$^{\dagger}$ & $-$1.33$^{\dagger}$ & 0.28$^{\dagger}$ & $-$0.01$^{\dagger}$ \\
\bottomrule
\end{tabular*}
\end{subtable}
\\[8pt]
\begin{subtable}[t]{0.485\textwidth}
\centering
\caption{Description (D)}
\begin{tabular*}{\linewidth}{@{\extracolsep{\fill}}lccccc@{}}
\toprule
\multirow{2}{*}{\textbf{Model}} & \textbf{2 Tools} & \multicolumn{2}{c}{\textbf{3 Tools}} & \multicolumn{2}{c}{\textbf{5 Tools}} \\
\cmidrule(lr){2-2}\cmidrule(lr){3-4}\cmidrule(lr){5-6}
 & \textbf{1H+1L} & \textbf{1H+2L} & \textbf{2H+1L} & \textbf{2H+3L} & \textbf{3H+2L} \\
\midrule
GPT-5.4 mini      & 0.22$^{\dagger}$ & 2.17 & 0.35$^{\dagger}$ & 1.15 & 0.06$^{\dagger}$ \\
Claude Sonnet 4.6 & 0.00$^{\dagger}$ & $-$0.02$^{\dagger}$ & 0.01$^{\dagger}$ & 0.24$^{\dagger}$ & $-$0.01$^{\dagger}$ \\
Gemini 3.5 Flash  & 1.59 & 3.50 & $-$1.83 & 2.12 & $-$0.55 \\
DeepSeek V4 Pro   & 5.97$^{\dagger}$ & 3.86 & 0.00$^{\dagger}$ & 1.43 & $-$0.37$^{\dagger}$ \\
Qwen3.7-Plus      & 0.00$^{\dagger}$ & sep$^{\ddagger}$ & sep$^{\ddagger}$ & 14.31$^{\dagger}$ & 13.72$^{\dagger}$ \\
\bottomrule
\end{tabular*}
\end{subtable}
\hfill
\begin{subtable}[t]{0.485\textwidth}
\centering
\caption{Parameter schema (P)}
\begin{tabular*}{\linewidth}{@{\extracolsep{\fill}}lccccc@{}}
\toprule
\multirow{2}{*}{\textbf{Model}} & \textbf{2 Tools} & \multicolumn{2}{c}{\textbf{3 Tools}} & \multicolumn{2}{c}{\textbf{5 Tools}} \\
\cmidrule(lr){2-2}\cmidrule(lr){3-4}\cmidrule(lr){5-6}
 & \textbf{1H+1L} & \textbf{1H+2L} & \textbf{2H+1L} & \textbf{2H+3L} & \textbf{3H+2L} \\
\midrule
GPT-5.4 mini      & 8.15$^{\dagger}$ & 4.04 & 4.50 & 5.02 & 5.60 \\
Claude Sonnet 4.6 & 7.79$^{\dagger}$ & 4.71 & 4.39 & 3.04 & 3.27 \\
Gemini 3.5 Flash  & 11.24$^{\dagger}$ & 14.03$^{\dagger}$ & 5.41 & 7.88 & 5.76 \\
DeepSeek V4 Pro   & 7.74$^{\dagger}$ & 3.85 & 1.90 & 4.05 & 3.26 \\
Qwen3.7-Plus      & 6.48$^{\dagger}$ & 12.76$^{\dagger}$ & 10.35$^{\dagger}$ & sep & sep \\
\bottomrule
\end{tabular*}
\end{subtable}
 
\vspace{4pt}
\begin{minipage}{\textwidth}
\scriptsize
\textit{Note.} Entries are conditional-logit coefficients ($\beta_f$) for $v_{\text{high}}$ vs.\ $v_{\text{low}}$, adjusted for position and estimated separately for each composition; $k$H+$l$L denotes $k$ candidates with $v_{\text{high}}$ and $l$ with $v_{\text{low}}$. The two-tool column repeats Table~\ref{tab:metadata-beta}. $^{\dagger}$95\% confidence interval includes zero; large estimates with $^{\dagger}$ reflect near-separation rather than weak effects. sep: separation, so no finite estimate exists. $^{\ddagger}$As in Table~\ref{tab:metadata_effects_three_tools}, Qwen's choices are determined by position, so these estimates should not be interpreted as effects.
\end{minipage}
\end{table}

\subsection{Parameter Names and Field Descriptions}
\label{app:naming_descriptions}
 
\begin{figure}[!htbp]
    \centering
    \includegraphics[width=\textwidth]{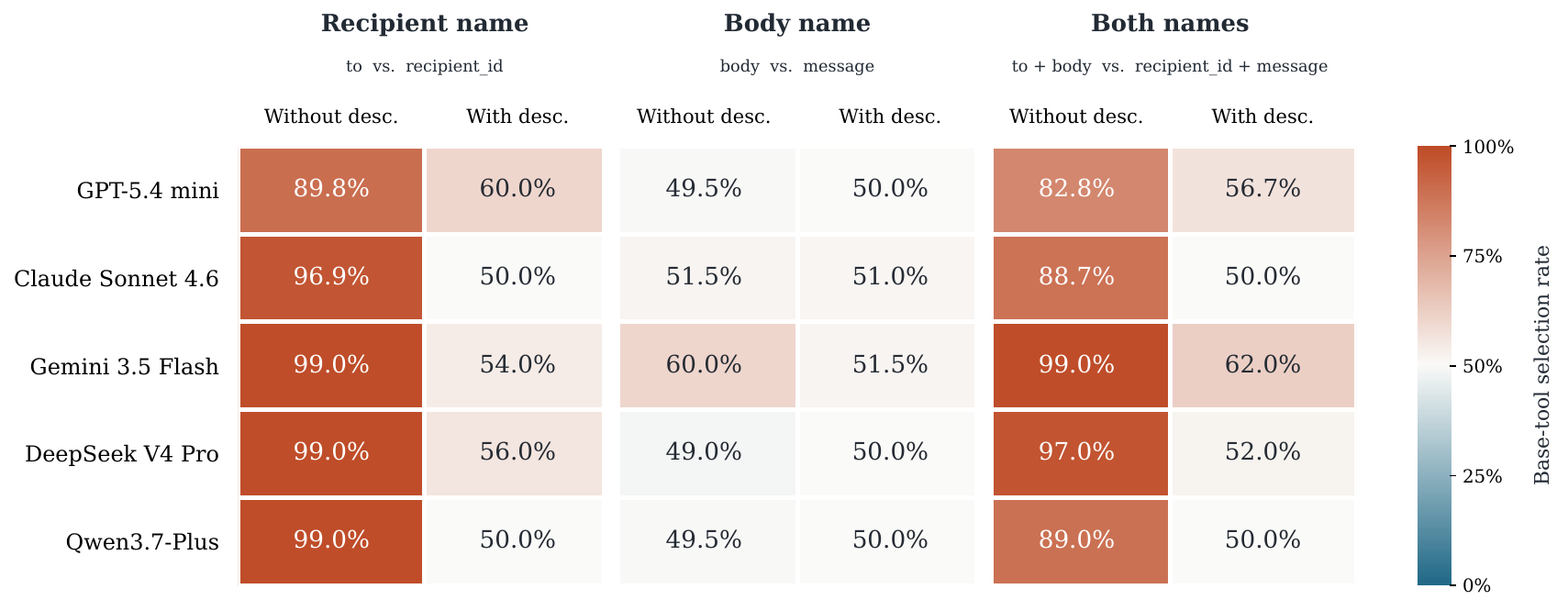} 
    \caption{Effects of parameter names and field descriptions on tool selection in Send Email. Cells show the percentage of valid trials in which the model selects the base tool, which uses \texttt{to}, \texttt{subject}, and \texttt{body}. The other tool renames \texttt{to} to \texttt{recipient\_id} (Recipient name), \texttt{body} to \texttt{message} (Body name), or both (Both names). \emph{With desc.}\ gives corresponding fields of both tools the same descriptions; \emph{Without desc.}\ omits field descriptions. The tool-level description is the same in all conditions, and 50\% indicates no preference.}
    \label{fig:naming_descriptions}
\end{figure}

Figure~\ref{fig:naming_descriptions} tests whether field descriptions change the naming effects in Section~\ref{sec:why}. In Send Email, the base tool uses \texttt{to}, \texttt{subject}, and \texttt{body}, and the other tool renames \texttt{to} to \texttt{recipient\_id}, \texttt{body} to \texttt{message}, or both. Each comparison is run without field descriptions and with the same descriptions on corresponding fields of both tools; all other metadata are identical. We use the 50 Send Email queries, with each tool in each position, giving 300 trials per model.
 
\paragraph{Recipient name.} Without field descriptions, all five models select the base tool in 89.8\%--99.0\% of trials, consistent with Section~\ref{sec:why}. With field descriptions, this preference largely disappears (50.0\%--60.0\%). The penalty for \texttt{recipient\_id} thus stems from how models interpret the name: once both fields carry the same description, the name no longer matters.
 
\paragraph{Body name.} Renaming \texttt{body} to \texttt{message} has little effect. Base-tool rates are 49.0\%--51.5\% without field descriptions, except 60.0\% for Gemini, and 50.0\%--51.5\% with them. Like \texttt{receiver} and \texttt{text}, \texttt{message} preserves the field's role.
 
\paragraph{Both names.} Renaming both fields yields base-tool rates of 82.8\%--99.0\% without field descriptions, no higher than for the recipient rename alone, so the effect is driven by \texttt{recipient\_id}. With field descriptions, the rates fall to 50.0\%--62.0\%.
 
Field descriptions therefore remove naming effects that arise from misleading parameter names, in contrast to their inconsistent effect on structured parameters in Place Order (Section~\ref{sec:why}).
 
\subsection{Task Requirements and Parameter Names}
\label{app:cc_naming}
 
\begin{figure}[!htbp]
    \centering
    \includegraphics[width=0.75\textwidth]{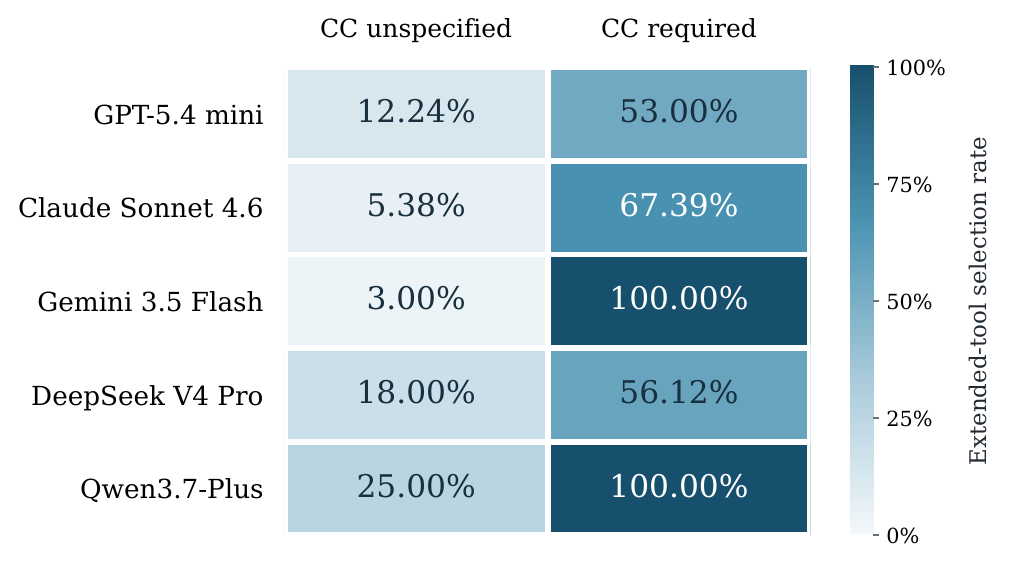} 
    \caption{Effects of explicit CC requests on tool selection in Send Email when the extended tool renames \texttt{to}. The base tool uses \texttt{to}, \texttt{subject}, and \texttt{body}; the extended tool uses \texttt{recipient\_id}, \texttt{subject}, \texttt{body}, and an optional \texttt{cc} field. Neither tool has field descriptions. Queries either do not mention CC (CC unspecified) or explicitly request CC recipients (CC required). Cells show the percentage of valid trials in which the model selects the extended tool. Valid trials contain exactly one call to a candidate tool, regardless of argument validity; each condition has 92--100 valid trials out of 100 (e.g., Qwen's 100\% under CC required is 97 of 97 valid trials).}
    \label{fig:cc_naming}
\end{figure}

Figure~\ref{fig:cc_naming} combines the task-requirement comparison of Section~\ref{sec:why} with the \texttt{recipient\_id} rename. The design follows Appendix~\ref{app:targeted}, except that the extended tool uses \texttt{recipient\_id} instead of \texttt{to}, so it both supports CC and carries the less preferred recipient name.
 
\paragraph{CC unspecified.} Without a CC request, the extended tool is selected in only 3.0\%--25.0\% of trials, compared with 45.0\%--63.1\% when it uses \texttt{to} (Section~\ref{sec:why}). The penalty for \texttt{recipient\_id} thus persists even when the tool offers an additional optional field.
 
\paragraph{CC required.} When the query explicitly requests CC recipients, selection of the extended tool rises for every model. Gemini and Qwen select it in every valid trial, but GPT, Claude, and DeepSeek do so in only 53.0\%--67.4\% of trials and otherwise choose the base tool, which has no \texttt{cc} field. By contrast, all five models select the extended tool in every valid trial when it uses \texttt{to} (Section~\ref{sec:why}). A misleading parameter name can therefore lead models to choose a tool that cannot fulfill an explicit request.
 
\section{Tool Selection in MCP}
\label{app:mcp_background}

\begin{figure}[h]
    \centering
    \includegraphics[width=\textwidth]{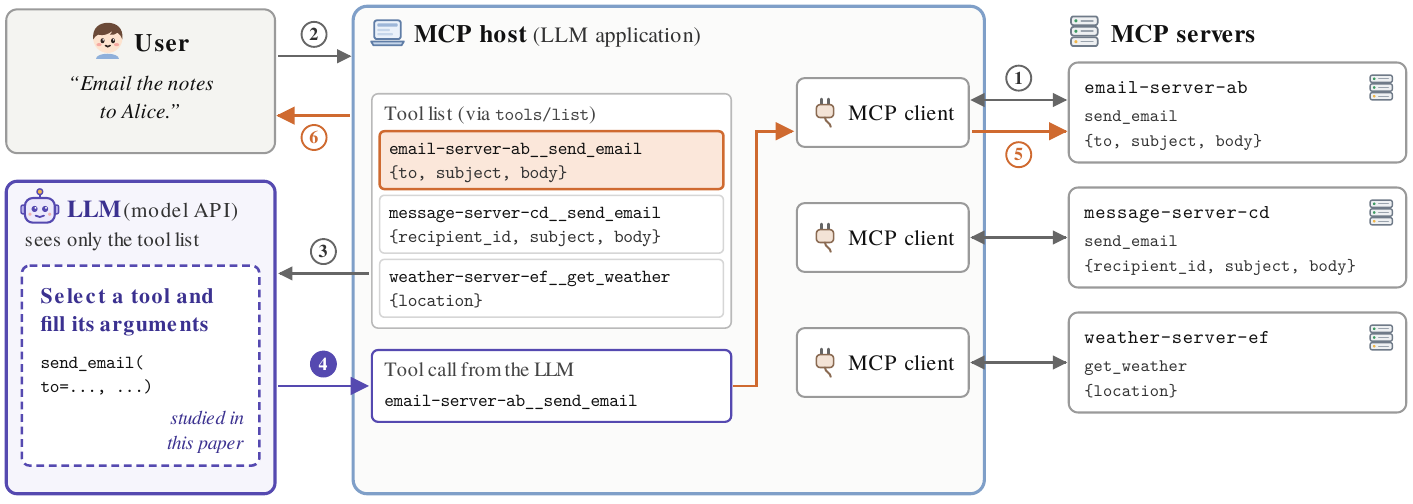}
    \caption{Tool selection in MCP. (1) For each connected server, the MCP host uses an MCP client that calls \texttt{tools/list} to obtain each tool's name, description, and input schema. (2) The user sends a request. (3) The host joins server and tool names into a single name field and sends the request with all tools to the LLM. (4) The LLM, which sees only this metadata, selects a tool, fills its arguments, and returns the tool call to the host; this is the step we study. (5) The host invokes the selected tool by sending \texttt{tools/call} to its server through the corresponding client, and (6) the result is returned to the user. Descriptions are omitted from the tool list for space.}
    \label{fig:mcp_workflow}
\end{figure}

This appendix outlines how an LLM selects tools in MCP~\citep{mcp_tools,mcp_client} and which part of this process our experiments examine. Figure~\ref{fig:mcp_workflow} illustrates the steps.

\paragraph{Components.}
An MCP \emph{host} is the LLM application with which the user interacts. The host connects to each MCP \emph{server} through its own MCP \emph{client}, and each server exposes one or more tools. The LLM is accessed by the host, typically through a model API, and does not communicate with clients or servers directly.
 
\paragraph{Discovery and presentation.}
Before a request is handled, each client calls \texttt{tools/list} on its server, which returns every tool's name, description, and input schema (step 1 in Figure~\ref{fig:mcp_workflow}). When the user sends a request (step 2), the host passes it to the LLM together with the metadata of all available tools (step 3). Because tools from different servers can share a name, the host must also indicate which server each tool belongs to; in our experiments, as in Figure~\ref{fig:mcp_workflow}, it joins the server and tool names into a single name field (e.g., \texttt{email-server-ab\_\_send\_email}). The LLM therefore sees exactly the four fields we manipulate: the server name ($S$) and tool name ($T$) in the joined name field, the description ($D$), and the parameter schema ($P$).
 
\paragraph{Selection.}
The LLM chooses one tool and fills its arguments based only on this metadata and the request, and returns the tool call to the host (step 4). It does not observe the tools' implementations, so functionally similar tools can be distinguished only by how their metadata is written. This selection step is the focus of our study.
 
\paragraph{Invocation.}
The host executes the tool call by sending \texttt{tools/call} with the chosen tool name and arguments to the corresponding server through its client (step 5). The result returns to the host, which delivers the final response to the user, typically after passing the result back to the LLM (step 6).
 
\paragraph{Scope of our experiments.}
Our experiments reproduce steps 3 and 4: each request presents the candidate tools' metadata to the LLM, and we record the tool it calls. No tool is executed, so our measurements reflect selection alone, independent of execution outcomes or feedback.

\end{document}